\documentclass{article}

    \PassOptionsToPackage{numbers,sort,compress}{natbib}

 \usepackage[final]{neurips_2020}

\usepackage[utf8]{inputenc} 
\usepackage[T1]{fontenc}    
\usepackage{hyperref}       
\usepackage{url}            
\usepackage{booktabs}       
\usepackage{amsfonts}       
\usepackage{nicefrac}       
\usepackage{microtype}      
\usepackage{wrapfig}

\usepackage{here}    
\usepackage{amsmath}
\usepackage{amssymb}
\usepackage{microtype}
\usepackage{graphicx}
\usepackage{booktabs} 
\usepackage{bm}
\usepackage{amsthm}
\newtheorem{thm}{Theorem}[section]
\newtheorem{lem}[thm]{Lemma}
\newtheorem{col}[thm]{Corollary}
\newtheorem{as}[thm]{Assumption}

\usepackage{color}
\usepackage{lipsum}

\usepackage{caption}
\usepackage{subcaption}

\newcommand{\com}[1]{\textcolor{black}{#1}}
\usepackage{wrapfig}

\title{Understanding Approximate Fisher Information for \\ Fast Convergence of Natural Gradient Descent \\ in Wide Neural Networks}

\author{%
 Ryo Karakida\\
  Artificial Intelligence Research Center\\
    AIST, Japan\\
    \texttt{karakida.ryo@aist.go.jp} \\
    \And
    Kazuki Osawa\\
    Department of Computer Science\\    
    Tokyo Institute of Technology, Japan\\
    \texttt{oosawa.k.ad@m.titech.ac.jp} 
}

\begin{document}

\maketitle

\begin{abstract}
Natural Gradient Descent (NGD) helps to accelerate the convergence of gradient descent dynamics, but it requires approximations in large-scale deep neural networks because of its high computational cost. Empirical studies have confirmed that some NGD methods with approximate Fisher information converge sufficiently fast in practice. Nevertheless, it remains unclear from the theoretical perspective why and under what conditions such heuristic approximations work well. In this work, we reveal that, under specific conditions, NGD with approximate Fisher information achieves the same fast convergence to global minima as exact NGD. We consider deep neural networks in the infinite-width limit, and analyze the asymptotic training dynamics of NGD in function space via the neural tangent kernel. In the function space, the training dynamics with the approximate Fisher information are identical to those with the exact Fisher information, and they converge quickly. The fast convergence holds in layer-wise approximations; for instance, in block diagonal approximation where each block corresponds to a layer as well as in block tri-diagonal and K-FAC approximations. We also find that a unit-wise approximation achieves the same fast convergence under some assumptions. All of these different approximations have an isotropic gradient in the function space, and this plays a fundamental role in achieving the same convergence properties in training. Thus, the current study gives a novel and unified theoretical foundation with which to understand NGD methods in deep learning.

\end{abstract}

\section{Introduction}
\label{Introduction}

Natural gradient descent (NGD) was developed to speed up the convergence of the gradient method \cite{amari1998natural}. The main drawback of the natural gradient is its high computational cost to compute the inverse of the Fisher information matrix (FIM). 
 Numerous studies have proposed approximation methods to reduce the computational cost so that NGD can be used in large-scale models with many parameters, especially, in deep neural networks (DNNs). For instance,  to compute the inverse efficiently, some studies have  proposed layer-wise block diagonal approximations of the FIM  \cite{heskes2000natural,martens2015optimizing,grosse2016kronecker}, where each block matrix corresponds to a layer of the DNN. This approach is usually combined with the Kronecker-factored approximate curvature (K-FAC) to further reduce the computational cost.   
Others have proposed unit-wise approximations, where each block matrix corresponds to a unit \cite{roux2008topmoumoute,ollivier2015riemannian,amari2018fisher}. 

Although empirical experiments have confirmed that these approximations make the convergence faster than the conventional first-order gradient descent, their conclusions are rather heuristic and present few theoretical guarantees on how fast the approximate NGD converges. It is important for both theory and practice to answer the question of how well approximate NGD preserves the performance of the original NGD.  
The lack of theoretical evidence is mostly caused by the difficulty of analyzing the training dynamics of neural networks. 
Recently, however, researchers have developed a theoretical framework, known as {\it neural tangent kernel (NTK)}, to analyze the training dynamics of the conventional gradient descent (GD) in DNNs with sufficiently large widths \cite{jacot2018neural,lee2019wide,arora2019exact}. The NTK determines the gradient dynamics in function space. 
It enables us to prove the global convergence of gradient descent and, furthermore, to explain generalization performance by using the equivalence between the trained model and a Gaussian process.

In this paper, we extend the asymptotic analysis of GD dynamics in infinitely-wide deep neural networks  to NGD and investigate the dynamics of NGD with the approximate FIMs developed in practice. We find that, surprisingly,  they achieve the same fast convergence of training to global minima as the NGD with the exact FIM. we show this is true for layer-wise block diagonal approximation of the FIM, block tri-diagonal approximation, K-FAC, and unit-wise approximation under specific conditions. Each algorithm requires an appropriately scaled learning rate depending on the network size or sample size for convergence. 
In function space, the exact NGD algorithm and these different approximations give the same dynamics on training samples. We clarify that they become independent of the NTK matrix and {\it isotropic in the function space}, which leads to fast convergence.

We also discuss some results with the goal of increasing our understanding of approximate NGD. First, the dynamics of approximate NGD methods on training samples are the same in the function space, but they are different in the parameter space and converge to different global minima. Their predictions on test samples also vary from one algorithm to another. Our numerical experiments demonstrate that the predictions of a model trained by the approximate methods are comparable to those of exact NGD. 
Second, we empirically show that the isotropic condition holds in the layer-wise and unit-wise approximations but not in entry-wise diagonal approximations of the FIM. 
In this way, we give a systematic understanding of NGD with approximate Fisher information for deep learning.

\section{Related work}

Although many studies have used NGD to train neural networks  \cite{amari1998natural,park2000,roux2008topmoumoute,ollivier2015riemannian,amari2018fisher,martens2015optimizing,pascanu2013,grosse2016kronecker,heskes2000natural}, 
our theoretical understanding of the convergence properties has remained limited to
shallow neural networks with a few units \cite{saad1998,cousseau2008dynamics} for decades. 
Moreover, although   
\citet{bernacchia2018exact} proved that NGD leads to exponentially fast convergence, this finding is limited to deep linear networks.
\citet{zhang2019fast} and \citet{cai2019gram} succeeded in proving the fast convergence of NGD in the NTK regime by using the framework of non-asymptotic analysis: they show a convergence rate better than that of GD \cite{zhang2019fast}, and quadratic convergence under a certain learning rate \cite{cai2019gram}.  
However, their analyses are limited to a training of the first layer of a shallow network. In contrast, we investigate NGD not only in shallow but also in deep neural networks, and derive its asymptotic dynamics; moreover, we consider the effect of layer-wise and unit-wise approximations on convergence.

Regarding the Fisher information, some studies have claimed that NGD with an empirical FIM (i.e., FIM computed on input samples $x$ and labels $y$ of training data) does not necessarily work well \cite{kunstner2019limitations,JMLR:v21:17-678}. 
As they recommend, we focus on NGD with a ``true'' FIM (i.e., FIM is obtained on input samples $x$ of training data,  and the output $y$ is analytically averaged over the true model) and its layer-wise and  approximations. Furthermore,
\citet{karakida2018universal} theoretically analyzed the eigenvalue spectrum of the FIM in deep neural networks on random initialization, but not the training dynamics of the gradient methods.

\section{Preliminaries}

\subsection{Gradient descent and NTK}
We focus on fully-connected neural networks: 
\begin{equation}
u_l=  \frac{\sigma_w}{\sqrt{M_{l-1}}}  W_{l} h_{l-1} +\sigma_b b_l, \ \ h_{l}= \phi(u_l), \label{eq1}
\end{equation}
for $l=1,...,L$, where we define activities $h_l \in \mathbb{R}^{M_l}$, weight matrices $W_l \in \mathbb{R}^{M_l \times M_{l-1}}$, bias terms $b_l \in \mathbb{R}^{M_l}$, and their variances $\sigma_w^2$ and $\sigma_b^2$. The width of the $l$-th layer is $M_l$, and we consider the limit of sufficiently large $M_l$ for hidden layers, i.e., $M_l=\alpha_lM$   and taking $M \gg 1$ ($\alpha_l>0$, $l=1,...,L-1$). We denote 
an input vector as $h_0=x$. The number of labels is given by a constant $M_L=C$. We suppose a locally Lipschitz and non-polynomial activation function $\phi(\cdot)$ whose first-order derivative $\phi'(\cdot)$ is also locally Lipschitz. 
Note that all of our assumptions are the same as in the conventional NTK theory of \cite{lee2019wide}. 
We consider random Gaussian initialization
\begin{equation}
    W_{l,ij}, b_{l,i} \sim \mathcal{N}(0,1),
\end{equation}
and focus on the mean squared error (MSE) loss 
\begin{equation}
    \mathcal{L}(\theta)= \frac{1}{2N}\sum_{n=1}^N \|y_n-f_\theta(x_n)\|^2,
\end{equation}
where the data samples ($x_n,y_n$) are composed of input samples  $x_n \in \mathbb{R}^{M_0}$ and labels $y_n \in \mathbb{R}^C$  ($n=1,...,N$). 
We normalize each sample so that $\|x_n\|_2=1$ and suppose $x_n \neq x_{n'}$ ($n \neq n'$) \cite{jacot2018neural,lee2019wide}.
The network model is given by $f_\theta= u_L$, and the set of all parameters is given by $\theta \in \mathbb{R}^P$. 

Here, we give an overview of the NTK theory of gradient descent (GD). 
The update rule of GD is given by 
\begin{equation}
    \theta_{t+1} = \theta_t - \eta \nabla_\theta   \mathcal{L}(\theta_t), \label{eq3:0326}
\end{equation}
where $\eta$ is a constant learning rate. 
The previous studies found that
 the dynamics of (\ref{eq3:0326}) in function space are asymptotically given by
\begin{equation}
 f_t(x') = \Theta(x',x)  \Theta(x,x)^{-1}(I-(I-\eta \Theta(x,x) )^t)(y-f_0(x))  +f_0(x'), \label{eq5:0326}
\end{equation}
 in the infinite-width limit of deep neural networks (\ref{eq1}) \cite{jacot2018neural,lee2019wide}. The notation is summarized as follows.  
We denote the identity matrix by $I$ and $f_{\theta_t}$ by $f_t$. Each  $f_t(x)$ and $y$ is a $CN$-dimensional vector which is the concatenation of  all $N$ samples.  
We denote the training input samples by $x$ and the arbitrary test samples by $x'$. When there are $N'$ test samples,  
$\Theta(x',x) $ is a $CN' \times CN$ matrix called as the neural tangent kernel:
\begin{equation}
\Theta(x',x) =  {J_0}(x') J_0(x)^\top/N,
\end{equation}
where $J_t(x)=\nabla_\theta f_t(x)$ is the $CN \times P$  Jacobian matrix.

  NTK dynamics (\ref{eq5:0326}) are interesting in the following points. First, the NTK defined at initialization determines the whole training process. This means that the dynamics (\ref{eq3:0326},\ref{eq5:0326}) are equivalent to those of a linearized model, i.e., $f_{t} = f_0 + J_{0} (\theta_t-\theta_0)$ \cite{lee2019wide}. Intuitively speaking,  we can train sufficiently wide neural networks in the range of a small perturbation around the initialization.  
  Second, as one can easily confirm by setting the training samples to $x'$, the training dynamics converge to $f_\infty(x)=y$. This means that the GD dynamics converge to a global minimum with zero training error in a sufficiently wide DNN. 
 The convergence speed is determined by the NTK, more precisely, by $(1-\eta \lambda_i)^t$, where the $\lambda_i$'s denote the NTK's eigenvalues. In general, in the linear model, convergence becomes slower as the eigenvalues become more distributed and the condition number becomes larger \cite{lecun1998efficient}. Finally,  $f_t(x')$ belongs to a Gaussian process. We can understand the generalization performance on the test samples $x'$ through Gaussian process regression \cite{jacot2018neural,lee2019wide}.

\subsection{NGD for over-parameterized models}
The natural gradient with a Riemannian metric of the parameter space $G$ \cite{amari1998natural} is given by $\theta_{t+1} = \theta_t - \eta \Delta \theta$, where
\begin{equation}
 \Delta \theta = G^{-1}_t \nabla_\theta   \mathcal{L}(\theta_t). \label{eq7:0523}
\end{equation}
The NGD for supervised learning with a mean squared error (MSE) loss has the following metric: 
\begin{equation}
    G_t =  F_t + \rho I, \ \ F_t:=  J_t^\top J_t/N.
\end{equation}
This is known as the Fisher information matrix (FIM) for MSE loss. 
In over-parameterized models, we add a non-negative damping term $\rho$ 
because $P>CN$ holds in most cases and $F_t$ is singular by definition. 
In particular, NGD with a zero damping limit  ($\rho \rightarrow 0$)  has a special meaning, as follows. 
For the MSE loss, we have $ \nabla_\theta   \mathcal{L} =  J^\top (f-y)/N$, and the natural gradient (\ref{eq7:0523}) becomes 
\begin{equation}
    \Delta \theta =  J_t^\top (J_t J_t^\top)^{-1} (f_t-y), \label{eq9:0524}
\end{equation}
where we have used the matrix formula $(J^\top J +\rho I)^{-1}J^\top = J^\top (J J^\top +\rho I)^{-1}$ \cite{cookbook} and take the zero damping limit.
This gradient is referred to as the NGD with the Moore-Penrose pseudo-inverse of $F_t$, which was first introduced by \cite{thomas2014genga} in a context different from neural networks and has recently been applied to neural networks \cite{bernacchia2018exact,zhang2019fast,cai2019gram}. Thus, the pseudo-inverse naturally appears in the NGD of over-parameterized models.
In the following analysis, we take the zero damping limit and use the pseudo-inverse in NGD of each approximate FIM. We call NGD (\ref{eq9:0524}) the exact pseudo-inverse NGD, or simply, the exact NGD.

\subsection{Overview of our formalization of NGD}

Before we show the details of the individual approximate methods, let us overview the direction of our analysis. In this study, we consider $G_t$ given by a certain approximate FIM. We show that, in the infinite-width limit, the dynamics of NGD (\ref{eq7:0523}) with the approximate FIM are asymptotically equivalent to
\begin{equation}
 f_t(x') = \bar{\Theta}(x',x)  \bar{\Theta}^{-1}(I-(I-\eta \bar{\Theta} )^t)(y-f_0) +f_0(x'). \label{eq61:0326}
\end{equation}
We leave the index of the test samples $x'$ and abbreviate the index of the training samples $x$ to $f_t=f_t(x)$ and $\bar{\Theta}=\bar{\Theta}(x,x)$ when the abbreviation  causes no confusion. 
We define the coefficient matrix of the dynamics by  \begin{equation}
\bar{\Theta}(x',x):= J_0(x')G_0^{-1} J_0(x)^\top/N.
\end{equation}
In the following sections, we show that various approximations to the FIM satisfy 
\begin{equation}
\bar{\Theta}(x,x) = \alpha I, \label{eq13:0410}
\end{equation}
on random initialization for a certain constant $\alpha>0$. We refer to this equation as the {\it isotropic condition}.
Under this condition, the NTK dynamics (\ref{eq61:0326}) become  $f_{t} = y+(1-\alpha \eta)^t (f_0-y)$ on the training samples $x$. 
All entries of the vector $f_{t}$ converge at the same speed $(1-\alpha \eta)^t$. This means that the isotropic condition makes the update in the function space isotropic. 
The training dynamics are independent of the NTK matrix, and the eigenvalue statistics of the NTK do not
slow down the convergence. In that sense, the dynamics of NGD (\ref{eq61:0326}) achieve fast convergence. In particular, if we set a learning rate satisfying $\eta=1/\alpha$, it converges in one iteration of training. This is reasonable since we suppose a quadratic loss and the model is asymptotically equal to the linearized model.

\noindent
{\bf Remark on exact NGD dynamics.} The NTK dynamics of NGD with the exact (pseudo-inverse) FIM (\ref{eq9:0524}) have been investigated in some previous studies \cite{zhang2019fast,rudner2019}. Assuming that the linearization of the DNN model in GD also holds in exact NGD, they showed that its NTK dynamics obey Eq. (\ref{eq61:0326}) with
\begin{equation}
\bar{\Theta}(x',x)=\Theta(x',x) \Theta^{-1}. \label{eq13:0524}
\end{equation}
Actually, we find that this linearization assumption is true in the infinite-width limit of deep neural networks. We give a proof in Section A of the Supplementary Material. 

Exact NGD accelerates the convergence of GD and converges to the same trained model, that is, 
\begin{equation}
f_\infty(x')= \Theta(x',x)\Theta^{-1}(y-f_0) + f_0(x'). 
\end{equation}
By substituting $f_t$ back into the update of $\theta_t$, we can confirm that GD and exact NGD reach the same global minimum:  $\theta_\infty-\theta_0 = J_0^\top \Theta^{-1}(y-f_0)/ N$. 
Similar to the case of GD \cite{jacot2018neural,lee2019wide}, we can interpret this prediction of the trained model as a kernel regression,  given by ${\Theta}(x',x)\Theta^{-1}y$ because the initialized model $f_0$ is a Gaussian process with zero mean.

\section{Layer-wise Fisher information}

In practice, we usually approximate the FIM to compute its inversion efficiently. A typical approach is to use block approximation where each block corresponds to a layer.  Block diagonal approximation uses only block diagonal matrices, and K-FAC further assumes a rough approximation of each diagonal block  \cite{heskes2000natural,martens2015optimizing,grosse2016kronecker}. We can also use tri-diagonal approximation, which includes interactions between neighboring layers, or even add higher-order interactions between distant layers.
In this section, we show that, under specific conditions, they achieve the same fast convergence as the exact NGD.

Before explaining the results of the individual layer-wise approximations, we show a general result for the layer-wise FIM. 
Consider the following class of layer-wise approximations:
\begin{equation}
    G_{\text{layer},t} :=\frac{1}{N} S_t^\top(\Sigma \otimes I_{CN})S_t + \rho I, \ \  S_t:=\begin{bmatrix}
 \nabla_{\theta_1} f_t           &     &         &  O \\
   &  \nabla_{\theta_2} f_t  &      &          \\
   &  & \ddots &   \\
 O  &        &       &  \nabla_{\theta_L} f_t      
 \end{bmatrix}. \label{eq15:0525}
\end{equation}
 $S_t$ is a $CNL \times P$ matrix whose diagonal block corresponds to a layer. We denote the set of parameters in the $l$-th layer by $\theta_l$, the Kronecker product by $\otimes$, and a $CN \times CN$ identity matrix by  $I_{CN}$. We suppose that $\Sigma \in \mathbb{R}^{L \times L}$ is a symmetric matrix and constant. For example, when $\Sigma$ is an identity matrix, 
$S^\top_t(\Sigma \otimes I_{CN})S_t$ becomes a block diagonal approximation to the FIM. The block tri-diagonal case corresponds to a specific $\Sigma$, shown in Section \ref{Sec4_2}. We compute the natural gradient by using the pseudo-inverse and set $\rho=0$.  

We obtain the following result: 
\begin{thm}[]
\label{thm4_1}
{\it Assume that $\Sigma$ is positive definite and define  $\Theta_l(x',x):= \nabla_{\theta_l} f_0(x') \nabla_{\theta_l} f_0(x)^\top/N$. For $0<\alpha\eta <2$, the dynamics of NGD with $G_{{\text{\rm  layer}},t}$  are asymptotically given by Eq. (\ref{eq61:0326}) with 
\begin{equation}
\bar{\Theta}(x',x)= \sum_{l=1}^L (\Sigma^{-1} 1_L)_l {\Theta}_l(x',x) {\Theta}_l^{-1}, \label{eq17:0504}
\end{equation}
in the infinite-width limit. 
The constant of the isotropic condition (\ref{eq13:0410}) is given by
 $\alpha = 1_L^\top \Sigma^{-1} 1_L$. 
}
\end{thm}
We denote  an $L$-dimensional vector all of whose entries are $1$ by $1_L$, and  the $i$-th entry of the vector $v$ by $(v)_i$. 
The derivation is given in the Supplementary Material. It is composed of three steps as shown in Section A: First, we prove that, under specific conditions (Conditions 1 and 2), 
NGD decreases the training loss to zero while keeping $\theta_t$ sufficiently close to $\theta_0$. Condition 1 is the isotropic condition, and Condition 2 is the local Lipschitzness of $G_t^{-1}J_t^\top$. Second, we prove that
the dynamics of approximate NGD is asymptotically equivalent to that of the linearized model, i.e., $f_{t} = f_0 + J_{0} (\theta_t-\theta_0)$.  These two steps of the proof is common among layer-wise and other approximations.
Finally,    
we show in Section B that Conditions 1 and 2 hold for layer-wise FIM.  After all, we obtain Eq. (\ref{eq61:0326}). 
 We can analytically compute each $\Theta_l(x',x)$ as shown in Section E. 
Regarding the learning rate, we have 
\begin{col}[]
\label{col4_2}
{\it   The dynamics of layer-wise NGD in Theorem 4.1 converge to the global minimum when   
\begin{equation}
    \eta= c/\alpha, \label{eq17:0530}
\end{equation}
where the constant is in the range $0<c<2$. In particular, given the optimal learning rate with $c=1$, the dynamics converge in one iteration of training.}
\end{col}
When $\eta=c/\alpha$, the training dynamics of layer-wise NGD
become $f_{t} = y+(1-c)^t (f_0-y)$. They are exactly the same as those of exact NGD with $\eta =c$.

The following sections describe the results of each approximate FIM.
 In addition to fast convergence on the training samples, the NTK dynamics (\ref{eq61:0326})  give some insight into generalization on test samples. Section \ref{Sec4_4} shows additional results on generalization.
 Although our analysis supposes the MSE loss, we can also give some insight into layer-wise FIMs for the cross-entropy loss. In the cross-entropy case, it is hard to obtain a closed form solution of the dynamics even under the assumption of linearization. Nevertheless, we can show that NGD with approximate FIMs  obeys the same update rule as that of the exact FIM (see Section D for the details).

\begin{figure}
\vspace{-5pt}
\centering
\begin{subfigure}[b]{.37\textwidth}
    \centering
    \includegraphics[width=\textwidth]{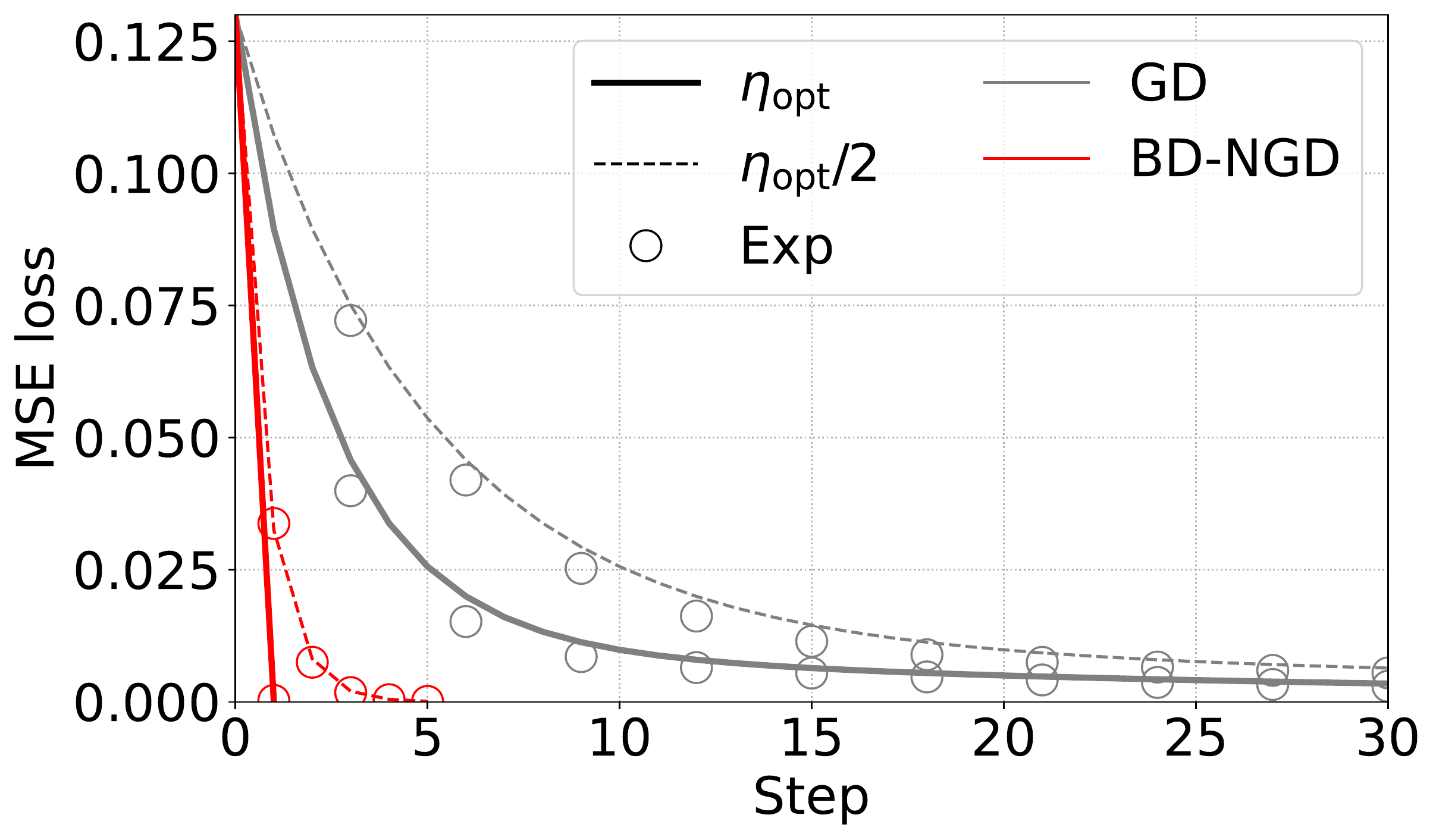}
    \caption{
    GD: $\eta_{\rm opt}=1/\lambda_{\rm max}(\Theta)$, BD-NGD: $\eta_{\rm opt}=1/L$; for networks with $L=3$.}
\end{subfigure}
\hfill
\begin{subfigure}[b]{0.6\textwidth}
    \centering
    \includegraphics[width=\textwidth]{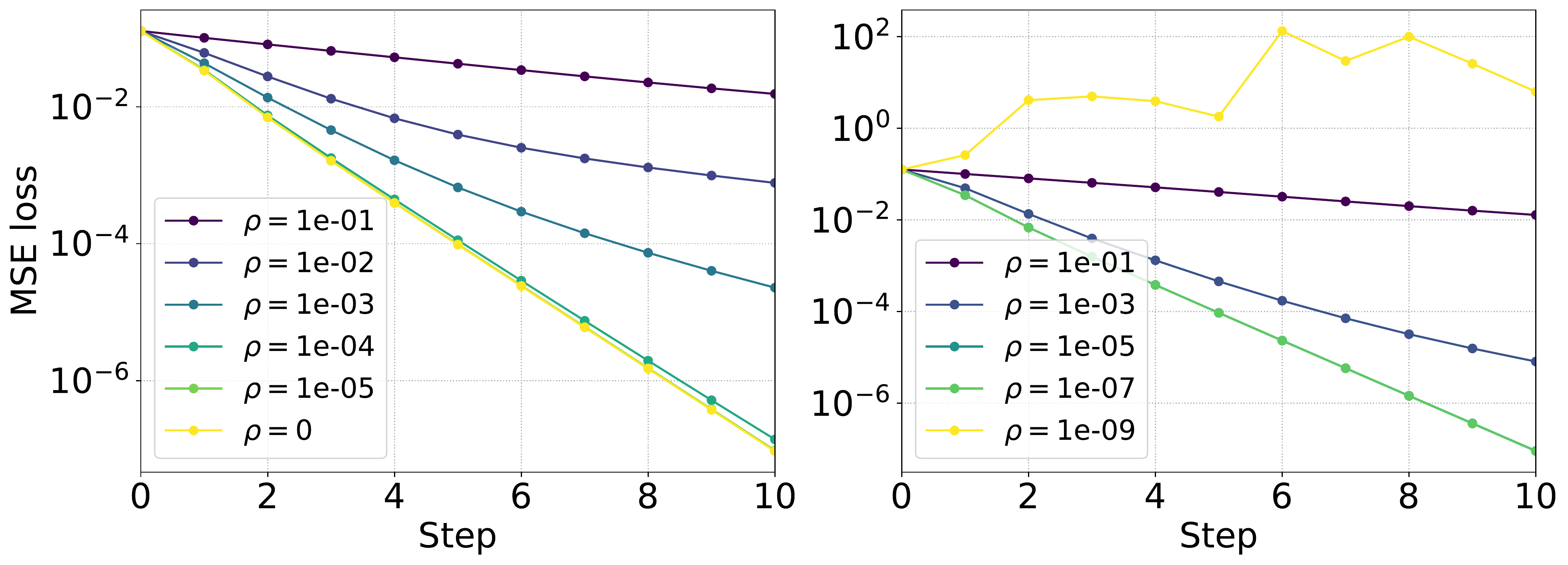}
    \caption{NGD with the block tri-diagonal FIM: $\eta=0.25$; for networks with $L=4$ (left), $L=5$ (right).
    }
\end{subfigure}
\caption{Fast convergence of NGD with layer-wise FIMs. Two-class classification on MNIST ('0' and '7') with deep ReLU networks, $N=100$, $M_l=4096$, $\sigma^2_w=2$, and $\sigma^2_b=0$.}
\vspace{-5pt}
\end{figure}

\subsection{Block-diagonal (BD) case}

This case corresponds to setting $\Sigma=I$. 
From Theorem \ref{thm4_1}, we immediately obtain
\begin{equation}
   f_t(x') = (1-(1-L\eta)^t)\frac{1}{L} \sum_{l=1}^L{\Theta}_l(x',x) {\Theta}_l^{-1}  (y-f_0(x)) +f_0(x') \label{eq18:0602}
\end{equation}
and $\alpha=L$. 
Despite that BD approximation neglects the non-diagonal blocks of the exact FIM, BD-NGD achieves the same convergence rate simply by setting a smaller learning rate scaled by $1/L$.  Figure 1(a) confirms that the training dynamics of numerical experiments (circles) coincide well with the NTK dynamics obtained by our theory (lines)\footnote{Source code is available at \url{https://github.com/kazukiosawa/ngd_in_wide_nn}.}.
We also plotted GD dynamics with an optimal learning rate 
$\eta_{\text{opt}}=1/\lambda_{max}(\Theta)$, which is recommended in \cite{lecun1998efficient} for fast convergence of GD. 
Even BD-NGD without its optimal learning rate converged faster than GD with its optimal learning rate.

\subsection{Block tri-diagonal case}
\label{Sec4_2}
Interestingly, we find that the convergence of the tri-diagonal case heavily depends on the depth $L$.  
The tri-diagonal approximation of the FIM is given by  a tri-diagonal matrix $\Sigma$,  
\begin{equation}
    \Sigma_{ij}=1 \ \ (i=j-1, j, j+1), \ \ 0 \ \ (\mathrm{otherwise}).
    \label{eq19:0607}
\end{equation}
The following lemma clarifies the dependence of the coefficient matrix $\bar{\Theta}$ on $L$:
\begin{lem}[]
\label{lem4_3}
{\it When $L=3s$ or $3s+1$  ($s=1,2 ...$), $\Sigma$ is positive definite and we have  $\alpha=s$ for $3s$ and  $\alpha=s+1$ for $3s+1$. 
In contrast, $\Sigma$ is singular when $L=3s+2$.
}
\end{lem}
The proof is given in Section B.2. 
Theorem 4.1 holds when $L=3s$ or $3s+1$. 
However, $\Sigma$ becomes singular and  the main assumption of  Theorem 4.1 does not hold when $L=3s+2$. Thus, we cannot guarantee the convergence of the training dynamics for this network. 
Figure 1(b) shows the results of numerical experiments on how the convergence depends on the depth and damping term. When $L=4$, the training dynamics got closer to that of $\rho=0$ (which is equal to the NTK dynamics (\ref{eq61:0326})) as the damping term decreased to zero. In contrast, when $L=5 \ (=3+2)$, the training dynamics exploded as the damping term became close to zero. This means that singularity of the block tri-diagonal FIM requires fine-tuning of the damping term for convergence. It is also hard to estimate the learning rate and damping term that give the fastest convergence.  

The dependence on the depth is in contrast to BD approximation, which holds for any depth. 
This suggests that adding higher-order interactions between different layers to the approximate FIM does not necessarily ensure the fast convergence of NGD.

\subsection{Kronecker-Factored Approximate Curvature (K-FAC)}

K-FAC is an efficient NGD algorithm for deep learning \cite{martens2015optimizing}. It supposes the BD approximation ($\Sigma=I$) and replaces the $l$-th layer's block by  
\begin{equation}
    G_{\text{K-FAC}} = (B_l^* + \rho I) \otimes (A_{l-1}^*+\rho I),
    \label{KFAC_F}
    \end{equation}
    where the Kronecker product reduces the computational cost of taking the inverse of the matrix. 
Matrices $A_l^*$ and $B_l^*$ come from feedforward signals and backpropagated signals, respectively. $A_l^*$  is given by a Gram matrix $h_l^\top h_l/N$, where $h_l \in \mathbb{R}^{N \times M_l}$ is a set of feedforward signals.
Let us denote the derivative by  
$\partial f(x_n)/ \partial W_{l,ij} = \delta_{l,i}(x_n) h_{l-1,j}(x_n)$.  $B_l^*$  is given by a Gram matrix $\delta_l^\top \delta_l/N$, where $\delta_l \in \mathbb{R}^{N \times M_l}$ denotes a set of backpropagated signals. 

For simplicity, we consider $C=1$, no bias terms, and $M_0 \geq N$. We also assume that input samples are linearly independent. Then, 
we find that the NTK dynamics are asymptotically given by Eq. (\ref{eq61:0326}) with   
\begin{equation}
\frac{1}{N}\bar{\Theta} (x',x)=\sum_{l=1}^{L-1}   (B_l(x',x)B_l^{-1}) \odot (A_{l-1}(x',x)A_{l-1}^{-1}) +  A_{L-1}(x',x)A_{L-1}^{-1},
\end{equation}
where $\odot$ means the Hadamard product and we define $A_l (x',x):=h_l(x')h_l(x)^\top/M_l$ and $B_l (x',x):=\delta_l(x')\delta_l(x)^\top$.  We can analytically compute the kernels $A_l$ and $B_l$ as is shown in Section E. 
Despite K-FAC heuristically replacing the diagonal block by the Kronecker product,  it satisfies the isotropic condition $\bar{\Theta}=NLI$. The optimal learning rate is given by $\eta_{\text{opt}}=1/(NL)$. The usual definition of K-FAC (\ref{KFAC_F}) includes an average over the training samples in both $A^*$ and $B^*$; it makes an extra $1/N$ in the function space and causes $\eta_{\text{opt}}$ to be proportional to $1/N$.   

We can generalize our result to the case of $M_0 < N$, where we have $\bar{\Theta}/N=(L-1)I + (I \odot X (X^\top   X)^{-1} X^\top)$. To achieve an isotropic gradient in this case, we need a pre-processing of input samples known as the Forster transformation. The necessity of this transformation was first reported by \cite{zhang2019fast}, who investigated the NTK of K-FAC in a shallow ReLU network without bias terms. We find that the Forster transformation is valid even in deep networks. It makes $X^\top   X \propto I $ and the isotropic condition holds.
We also find that K-FAC achieves the fast convergence in networks with bias terms. It remains for future research to investigate $C>1$. 
The details are shown in Section B.3.

\begin{figure}
\vspace{-5pt}
\centering
\includegraphics[width=0.85\textwidth]{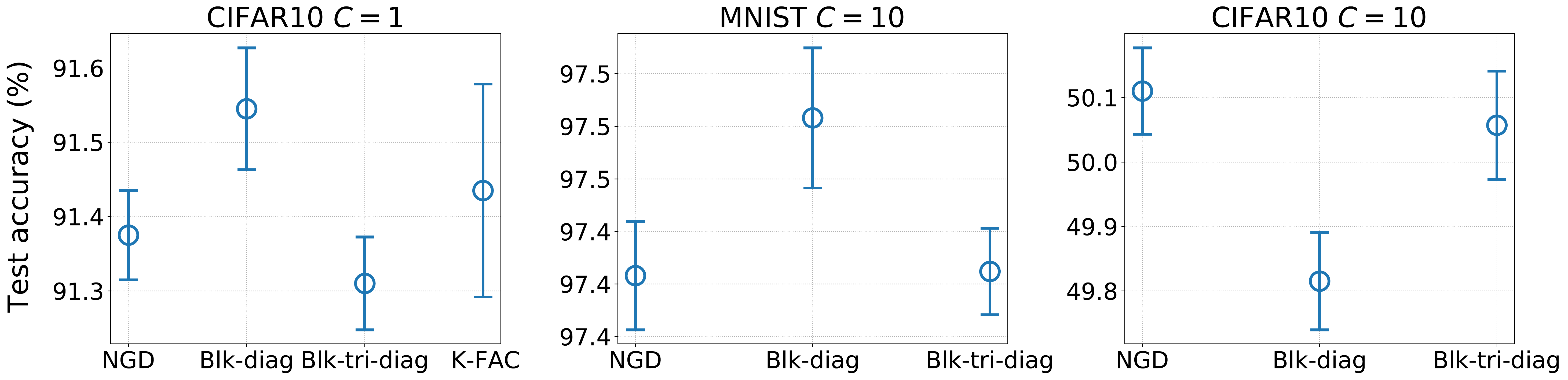}
\caption{Prediction of the trained model. We set $L=3$, $M_l=4096$, $\sigma_w^2=2$, and $\sigma_b^2=0$.
Left: two-class (airplane and horse), $N=10000$, $N'=2000$. Center and right: ten-class, $N=10000$, $N'=10000$. The mean and standard deviation are calculated from 10 independent initializations.}
\vspace{-5pt}
\end{figure}

\subsection{Points of difference among approximate FIMs}
\label{Sec4_4}
In the above sections, we found that the layer-wise FIMs show essentially the same convergence properties in training in the function space. This raises a natural question as to whether these approximation methods have differences in any other aspects. Actually, each approximation has its own implicit bias in the following two points.     

\noindent
{\bf Solution in the parameter space:}
 We can also obtain the dynamics of the parameter $\theta_t$ by substituting the obtained dynamics in the function space $f_t$ back into the update in the parameter space. The training dynamics in the function space are essentially the same among the different approximations, but $\theta_t$ is different; 
\begin{equation}
    \theta_{t} -\theta_0 =\alpha^{-1} (1-(1-\alpha \eta)^{t}) G_0^{-1} J_0^\top  (y-f_0)/N.
\end{equation}
For instance, we have $G_0^{-1} J_0^\top/N= S_0^\top (S_0S_0^\top)^{-1} (\Sigma^{-1}1_L \otimes I)$ for layer-wise approximations (\ref{eq15:0525}).
An over-parameterized model has many global minima, and each algorithm  chooses a different minimum depending on the approximation that it uses.  All these minima can be regarded as min-norm solutions with different distance measures. 
Taking the average over the random initializations, we have $\theta_{\infty}= \alpha^{-1} G_0^{-1} J_0^\top y/N$; this is equivalent to
a min-norm solution $\mathrm{argmin}_\theta \frac{1}{2N} \|y-  J_0\theta\|^2_2+ \frac{\lambda}{2} \theta^\top G_0 \theta$ in the ridge-less limit ($\lambda \rightarrow 0$). 
The derivation is given in Section B.4.

\noindent
{\bf Prediction on test samples: }
Although our main purpose is to understand convergence in training, we can also give insight into prediction. In the same way as GD, we can interpret the trained model as a kernel regression, that is, $\alpha^{-1} \bar{\Theta}(x',x)y$. The  matrix $\bar{\Theta}(x',x)$ and the predictions on the test samples vary depending on the approximations used. For instance, the prediction of the BD approximation is given by $\sum_l L^{-1}\Theta_l(x',x)\Theta_l^{-1} y$. This means that the model trained by BD-NGD can be regarded as an average over the estimators obtained by training each layer independently. Moreover, one can view the tri-diagonal case (\ref{eq17:0504}) as a modification of BD weighted by $(\Sigma^{-1} 1_L)_l$.

Figure 2 shows the results of numerical experiments with deep ReLU networks on the MNIST and CIFAR-10 datasets. We calculated the test accuracy by using $f_\infty(x')$ for each $\bar{\Theta}(x',x)$: exact NGD, BD-NGD, block tri-diagonal NGD, and K-FAC (only for $C=1$).
As is summarized in Section E, we used the analytical representations of $\bar{\Theta}(x',x)$.
Each circle corresponds to $\alpha^{-1} \bar{\Theta}(x',x)y$. Note that the variance appears because $f_0$ is a Gaussian process depending on the random initialization. 
We can see that the test accuracy varies depending on the approximate FIMs used, but are comparable to each other. Since the performance also depends on the data, it is hard to choose which FIM is generally better.
This suggests that the model trained by approximate NGD has sufficient performance.

\section{Unit-wise Fisher information}

We consider a unit-wise block diagonal approximation of the FIM: 
\begin{equation}
    G_{\text{unit},t} := \frac{1}{N} S_{\text{unit},t}^\top  S_{\text{unit},t}+ \rho I, \label{eq24:0602}
\end{equation}
where $S_{\text{unit},t}$ is a  $CN(\sum_{l=1}^L M_l) \times P$ block diagonal matrix whose $j$-th block corresponds to the $i$-th unit in the $l$-th layer,  i.e.,  $ \nabla_{\theta_i^{(l)}} f_t $ ($j=i+\sum_{k=1}^{l-1}M_k$).
We denote the set of parameters in the unit by $\theta_i^{(l)}=\{W_{l,i1},...,W_{l,iM_{l-1}},b_{l,i} \}$. Then, the $j$-th block of $S_{\text{unit}}^\top  S_{\text{unit}}$ is  $\nabla_{\theta_i^{(l)}} f_t^\top \nabla_{\theta_i^{(l)}} f_t$. Note that we take the pseudo-inverse and zero damping limit for the computation of the natural gradient. This naive implementation of the  unit-wise NGD requires roughly $LM$ $M\times M$ matrices to be stored and inverted, while K-FAC only requires $2L$ $M\times M$ matrices. Although   some studies on unit-wise NGD further approximated the unit-wise FIM  (\ref{eq24:0602}) and proposed more efficient algorithms for practical use \cite{roux2008topmoumoute,ollivier2015riemannian,amari2018fisher}, we focus on the naive implementation of the unit-wise NGD as a first step.

For simplicity, we consider $C=1$, $M_l=M$, $M_0\geq N$ and assume that input samples are linearly independent. 
In addition, we require {\it the gradient independence assumption} which is commonly used in the mean field theory of DNNs  \cite{schoenholz2016,yang2017,xiao2018dynamical,karakida2019normalization}. 
That is, in the computation of backpropagated gradients ($\delta_l$) on random initialization, we replace the transposed weight $W_l^\top$ by a fresh i.i.d. copy $\tilde{W}_l$. 
We use this assumption for proving the isotropic condition\footnote{\citet{yang2020tensor} has recently proved that when the activation function is polynomially bounded, using the gradient independence assumption leads to  correct results. This justification is applicable to our Theorem 5.1.}, which  includes a summation of $\delta_{l,i}$ over units and this is quite similar to the derivation of order parameters in the mean field theory. 
 We find that the fast convergence holds on the training samples (see Section C for the proof): 
\begin{thm}[]
\label{thm5_1}
{\it Under the gradient independence assumption and for the zero damping limit $ \rho=1/M^\varepsilon$ ($0< \epsilon< 1/12$), the training dynamics of NGD with $G_{{\text{\rm  unit}},t}$ are asymptotically given by 
\begin{equation}
f_{t} = (1-(1-\alpha \eta )^t)(y-f_0) +f_0, \ \ \alpha= \gamma M(L-1), \label{eq24:0601}
\end{equation}
 in the infinite-width limit, where $\gamma$ is a positive constant. To make the training converge, we need a learning rate $\eta=c/\alpha$ ($0<c<2$), and the optimal learning rate is $c=1$.
}
\end{thm}
As is shown in the proof, $\gamma$ depends on the shape of the activation function. For instance, we have $\gamma=1$ for Tanh and $\gamma=1/2$ for ReLU activation. 
 Although the current proof approach requires the assumption, 
 we confirmed that the obtained training dynamics coincided well with experimental results of training (see Section C.3). 
The unit-wise approximation uses only $1/(M(L-1))$ entries of the exact FIM, and it is much smaller than the exact and layer-wise FIMs (where we measure the size by the number of non-zero entries of the matrix). Nevertheless, the unit-wise NGD can converge with the same rate of convergence as the exact NGD. We also derive corresponding results for $M_0<N$. In this case, the isometric condition holds when parameters in the first layer are fixed.

\begin{wrapfigure}{r}{0.33\textwidth}
  \vspace{-14pt}
  \begin{center}
    \includegraphics[width=0.33\textwidth]{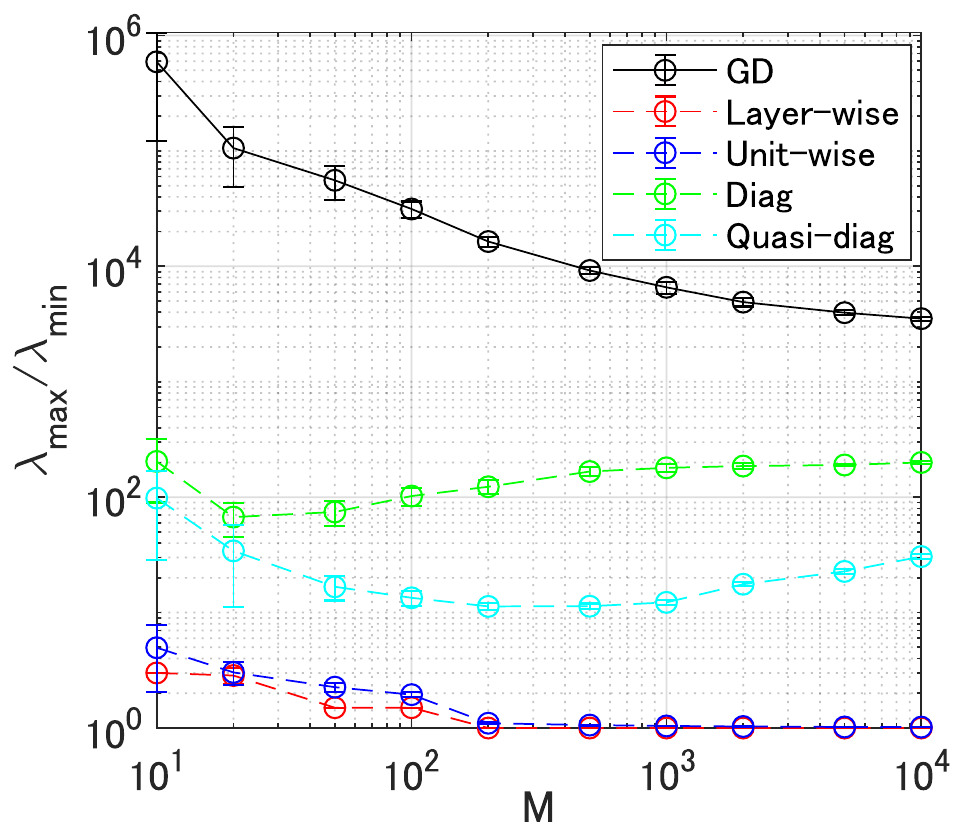}
  \end{center}
  \vspace{-1pt}
  \caption{Condition number of $\bar{\Theta}$}
   \vspace{-3pt}
\end{wrapfigure}
\noindent
{\bf Comparison with entry-wise FIMs.} 
Figure 3 shows the results of numerical experiments on the isotropic condition (\ref{eq13:0410}). We used a ReLU network with $L=3$ and Gaussian inputs (see Section C.3 for more details).
We computed the eigenvalues of $\bar{\Theta}$ on random initialization and measured the degree of isotropy in terms of condition number ($:=\lambda_{max}/\lambda_{min}$). When the condition number takes $1$, all eigenvalues take the same value and the isotropic condition holds. As we expect, the condition numbers of $\bar{\Theta}$  in BD-NGD (red circles) and in unit-wise NGD (blue circles) took $1$ in large widths.
For comparison, we also show the condition numbers of NTK (black circles),  $\bar{\Theta}$ with an entry-wise diagonal FIM (i.e., $G_{ij}=(F_{ii}+\rho)\delta_{ij}$) \cite{lecun1998efficient} (green circles), and  $\bar{\Theta}$ with the quasi-diagonal FIM \cite{ollivier2015riemannian} (cyan circles). The quasi-diagonal FIM was proposed as a rough approximation of the unit-wise FIM in which a certain 1-rank matrix is added to the diagonal entries  \cite{ollivier2015riemannian,amari2018fisher}. 
We find that these entry-wise FIMs had better condition numbers than NTK, but they kept taking larger values than $1$ even in the case of a large width and they did not satisfy the isotropic condition. This suggests that NGD with entry-wise approximations will converge faster than GD but not than layer-wise and unit-wise ones.
It would be interesting to explore any approximation satisfying the isotropic condition that is larger than the entry-wise approximation but smaller than the unit-wise one.

\section{Conclusion and future directions}
We provided a unified theoretical backing on natural gradient with various approximate FIMs. 
Through the lens of NTK, we found that they achieve the same fast convergence as the exact natural gradient under specific conditions. 
Despite that the approximate FIMs are different from each other, they share the same isotropic gradient in function space.

While the main purpose of the current work is to achieve a theoretical understanding of the NGD dynamics, it is also  important to develop efficient NGD algorithms with low computational complexity. It would be interesting to explore NGD algorithms satisfying the isotropic condition and keeping the computational cost as low as possible. 
To further increase the scope of our theory,
it would be interesting to investigate NGD in convolutional neural networks \cite{grosse2016kronecker} by leveraging the NTK theory developed for them \cite{arora2019exact}. 
Developing a non-asymptotic analysis of NGD will also be helpful in quantifying the effect of a finite width on the convergence. 
We expect that as the theory of NTK is extended into more various settings, it will further shed light on the design of natural gradient algorithms in deep learning.

\section*{Broader Impact}
We believe that this section is not applicable to this paper.

\section*{Acknowledgements}
We thank the reviewers for insightful and helpful reviews of the manuscript. We also thank Shun-ichi Amari for his insightful comments, and the members of ML Research Team in AIST for their useful discussions.  
RK acknowledges the funding support from JST ACT-X Grant Number JPMJAX190A. KO is a Research Fellow of JSPS and is supported by JSPS KAKENHI Grant
Number JP19J13477.

\nocite{yang2020tensor}
\nocite{willett1965discrete}
\nocite{noschese2013tridiagonal}
\nocite{da2001explicit}

\bibliographystyle{unsrtnat}
\bibliography{test}

\newpage

\setcounter{section}{0}
\renewcommand{\thesection}{\Alph{section}}
\renewcommand{\theequation}{S.\arabic{equation}}
\setcounter{equation}{0}

\part*{Supplementary Materials}

\section{NTK dynamics of NGD: General formulation}
The proofs for the convergence of NGD dynamics share a common part among various types of approximations. Therefore, we first introduce specific conditions that are necessary to prove the convergence (Conditions 1 and 2), and reveal the convergence under these conditions (Theorem A.3). Later, 
we prove that each approximate FIM satisfies Conditions 1 and 2 (layer-wise FIMs in Section B and unit-wise FIM in Section C). 

As preparation for analysis, we summarize our assumptions mentioned in the main text;

\noindent
{\bf Assumption 1.}
{\it  The activation function $\phi(\cdot)$ is locally Lipschitz and grows non-polynomially. Its first-order derivative $\phi'(\cdot)$ is also locally Lipschitz. }

\noindent
{\bf Assumption 2.}
{\it  Suppose training samples normalized by $\|x_n\|_2=1$, and $x_n \neq x_{n'}$ $(n\neq n')$. }

These assumptions are the same as in the NTK theory for GD \cite{jacot2018neural,lee2019wide}. 
 Assumption 2 is used to guarantee the positive definiteness of NTK or its variants.  
Assumption 1 plays an essential role in the conventional theory of GD  through the following Lemma. 
\begin{lem}[\cite{lee2019wide}; {\bf Local Lipschitzness of the Jacobian}] 
Assume Assumption 1. There is a constant $K>0$ such that for a sufficiently large $M$ and every $D>0$,
  with high probability (w.h.p.) over random initialization we have 
  \begin{align}
&M^{-\frac{1}{2}}\|h_{l}(\theta)\|_{2},  \quad\|\delta_{l}(\theta)\|_{2}
 \leq K,  \\ 
&M^{-\frac{1}{2}}\|h_{l}(\theta)-h_{l}(\tilde{\theta})\|_{2}, \quad\|\delta_{l}(\theta)-\delta_{l}(\tilde{\theta})\|_{2}\leq  K\|\tilde{\theta}-\theta\|_{2}/\sqrt{M},
  \end{align}
  and 
\begin{equation}
    \left\{\begin{array}{ll}
 \|J(\theta)\|_{F} & \leq K,\\
\|J(\theta)-J(\tilde{\theta})\|_{F} & \leq K\|\theta-\tilde{\theta}\|_{2}/\sqrt{M} 
\end{array} \quad \quad \forall \theta, \tilde{\theta} \in B\left(\theta_{0}, D \right),\right. \label{S9:0609}
\end{equation}  
  where a ball around the initialization is defined by $B\left(\theta_{0}, D\right):=\left\{\theta:\left\|\theta-\theta_{0}\right\|_{2}<D\right\}$.
\end{lem}
The constants $K$ and $D$ may depend on $\sigma_w^2$, $\sigma_b^2$, $N$ and $L$, but independent of $M$. The matrix norm $||\cdot||_F$ denotes the Frobenius norm.
The meaning of w.h.p. is that the proposition holds with probability 1 in the limit of large $M$.

Note that we adopt the NTK parameterization as is usual in the studies of NTK \cite{jacot2018neural,lee2019wide,arora2019exact}. 
That is, we initialize $W$ by a normal distribution with a variance 1, and normalize $W$ by the coefficient $1/\sqrt{M}$ in Eq. (1). 
In contrast, parameterization defined by $\theta' = \{W', b'\}$
 with $W' \sim \mathcal{N}(0, \sigma_w^2/M)$ and  $b' \sim  \mathcal{N}(0, \sigma_b^2)$ is so-called the standard parameterization.
 NTK dynamics in the NTK parameterization with a constant learning rate $\eta$ is equivalent to that in the standard parameterization with a learning rate $\eta/M$ \cite{lee2019wide}. 

We denote the coefficient of the dynamics at time step $t$ by 
\begin{equation}
    \bar{\Theta}_t(x',x) := J_t(x') G_t(x)^{-1} J_t(x)^\top/N,
\end{equation}
where  $G_t(x)$ is the FIM on the training samples. 
We represent $\bar{\Theta}_0(x',x)$ by $\bar{\Theta}(x',x)$, and $\bar{\Theta}(x,x)$ by $\bar{\Theta}$ on training samples $x$, if such abbreviation causes no confusion. 
Now, we introduce two conditions to be satisfied by approximate FIMs.  

\noindent
{\bf Condition 1} ({\bf Isotropic Condition}){\bf.} {\it on random initialization, the following holds }
\begin{equation}
    \bar{\Theta}=\alpha I.
\end{equation}

{\bf Condition 2.}
{\it There is a constant $A>0$ such that for a sufficiently large $M$ and every $D>0$, with high probability, the following holds }
\begin{equation}
 \left\{\begin{array}{ll}
    \bar{\eta} \| G_s^{-1} J_s^\top \|_{{2} } &\leq A, \\
    \bar{\eta} \| G_0^{-1}J_0^\top -G_s^{-1} J_s^\top \|_{{2} } &\leq A\|\theta_s-\theta_0\|_2 /\sqrt{M}
\end{array} \quad \quad \forall \theta_s \in B\left(\theta_{0}, D \right).\right.
\end{equation}
We define a scaled learning rate $\bar{\eta}= \eta/N$.  The matrix norm $||\cdot||_2$ denotes the spectral norm.
Condition 2 is a counterpart of the Lipschitzness of the Jacobian (\ref{S9:0609}) in GD. 
We denote $\tilde{\theta}$ by $\theta_s$, and  $J(\theta_s)$ by $J_s$. This notation is intuitive because  we prove Theorem A.2 by induction on the parameter $\theta_t$ at time step $t$ and use Condition 2 at each induction step.  
We show later that these conditions hold for our approximate FIMs.

\subsection{Global convergence around the initialization}

The proof is composed of two parts.  First, we show that the training loss monotonically decreases to zero (Theorem A.2).  Second, we use Theorem A.2 and prove that NGD dynamics of wide neural networks are asymptotically equivalent to those of linearized models (Theorem A.3). This approach is similar to the previous work on GD \cite{lee2019wide}.

Let us denote the training error by $g(\theta_t):=f_t-y$. We have the following.  
\begin{thm}
\label{ThmA2}
Assume   Assumptions 1 and 2, and that Conditions 1 and 2 hold.  For $0<\eta \alpha<2$ and a sufficiently large $M$, the following holds with high probability,
\begin{align}
\|g (\theta_{t} )\|_{2} &\leq \left(|1- \eta \alpha |+\frac{A'}{\sqrt{M}}\right)^{t} R_{0}, \label{S5:0527} \\
 \sum_{j=1}^{t}\|\theta_{j}-\theta_{j-1}\|_{2} &\leq AR_0\sum_{j=1}^{t} \left(|1- \eta \alpha |+\frac{A'}{\sqrt{M}}\right)^{j-1} \leq \frac{2AR_0}{1-|1-\eta \alpha|}, \label{S11:0531}
\end{align}
with $A'= 4KA^2 R_0/(1-|1-\eta \alpha|)$.
\end{thm}

\noindent
{\it Proof.}
We prove the inequalities (\ref{S5:0527}, \ref{S11:0531}) by induction.  It is obvious that we have  
\begin{equation}
    \|g(\theta_0)\|_2 < R_0.
\end{equation}
for a constant $R_0>0$ \cite{lee2019wide}. It is easy to see that the inequality (\ref{S5:0527}) holds for $t=0$ and (\ref{S11:0531}) hold for $t=1$. Suppose that the inequalities (\ref{S5:0527},\ref{S11:0531})  holds at a time step $t$. Then, we prove the case of $t+1$ as follows. First, 
note that we have $|1- \eta \alpha|<1$ and 
\begin{equation}
    \|\theta_{t+1}-\theta_t \|_2 \leq \bar{\eta}  \|G^{-1}_t J_{t} \|_{{2} } \| g(\theta_t) \|_2 \leq  A R_0 \left(|1- \eta \alpha |+\frac{A'}{\sqrt{M}}\right)^{t}. 
\end{equation} 
For a sufficiently large $M$, $|1- \eta \alpha |+\frac{A'}{\sqrt{M}}<1$ holds and we obtain the desired inequality (\ref{S11:0531}). Next, 
The error at $t+1$ is given by
\begin{align}
\|g\left(\theta_{t+1}\right)\|_{2} &=\|g\left(\theta_{t+1}\right)-g\left(\theta_{t}\right)+g\left(\theta_{t}\right)\|_{2} \\
&=\|\tilde{J}_t\left(\theta_{t+1}-\theta_{t}\right)+g\left(\theta_{t}\right)\|_{2} \label{eqMVT} \\
&=\|-\bar{\eta} \tilde{J}_t G_t^{-1} J\left(\theta_{t}\right)^\top g\left(\theta_{t}\right)+g\left(\theta_{t}\right)\|_{2} \\
& \leq \|I-\bar{\eta} \tilde{J}_t G_t^{-1} J(\theta_{t})^\top \|_{{2} }\left\|g\left(\theta_{t}\right)\right\|_{2} \\
& \leq \|I-\bar{\eta} \tilde{J}_{t} G_t^{-1} J(\theta_{t})^\top \|_{{2} }\left(|1- \eta \alpha |+\frac{A'}{\sqrt{M}}\right)^{t} R_{0}, \label{S14:0611}
\end{align}
where we define $\tilde{J}_t = \int_0^1 J( \theta_t + s(\theta_{t+1}- \theta_t) )ds$.
Here,
\begin{align}
\|I-\bar{\eta} \tilde{J}_{t} G_t^{-1} J(\theta_{t})^\top \|_{{2} } & \leq \| I- \eta \bar{\Theta} \|_{{2} } +\eta \|\bar{\Theta} -\tilde{J}_{t} G_t^{-1} J(\theta_{t})^\top/N\|_{{2} }. \label{S20:0608}
\end{align}
Using Condition 1, we have 
\begin{equation}
    \| I- \eta \bar{\Theta} \|_{{2} } = |1-\eta \alpha|.
\end{equation}
In addition, we have
\begin{align}
&\eta \|\bar{\Theta} -\tilde{J}_{t} G_t^{-1} J(\theta_{t})^\top/N\|_{{2} } \label{S19:0609} \nonumber \\
&\leq \bar{\eta} \|J_{0}G_0^{-1} J_{0}^\top - J_{0}G_t^{-1} J_{t}^\top \|_{{2} } + \bar{\eta} \|J_{0}G_t^{-1} J_{t}^\top -\tilde{J}_{t}G_t^{-1} J_{t}^\top\|_{{2} } \\ 
&\leq \bar{\eta} \|G_0^{-1} J_{0}^\top - G_t^{-1} J_{t}^\top \|_{{2} } \|J_{0} \|_{{2} }  + \bar{\eta}  \|G_t^{-1} J_{t}^\top\|_{{2} }  \|J_{0}-\tilde{J}_{t}\|_{{2} },\label{S19:0000}
\end{align}
and
\begin{align}
    \|J_{0}-\tilde{J}_{t}\|_{{2} } & \leq \int_0^1 \|J_{0}-J( \theta_t + s(\theta_{t+1}- \theta_t))\|_{{2} }  ds \\ 
    & \leq K( \| \theta_t- \theta_0 \|_2 + \| \theta_{t+1}- \theta_t \|_2)/\sqrt{M}. \label{S21:0000}
 \end{align}
Then, using  Condition 2 in (\ref{S19:0000}) and (\ref{S11:0531})  in (\ref{S21:0000}), we obtain
\begin{align}
 \eta \|\bar{\Theta} -\tilde{J}_{t} G_t^{-1} J(\theta_{t})^\top/N\|_{{2} }  
&\leq A'/\sqrt{M},\label{S20:0929}
\end{align}
Substituting (\ref{S20:0608})-(\ref{S20:0929}) into (\ref{S14:0611}), we have 
\begin{equation}
    \|g\left(\theta_{t+1}\right)\|_{2} \leq \left(|1- \eta \alpha |+\frac{A'}{\sqrt{M}}\right)^{t+1}R_0.
\end{equation}

\qed

\subsection{Bounding the discrepancy between the original and the linearized model}

Let us consider a linearized model given by 
\begin{equation}
  f_t^{{lin}}(x) := f_0(x)+J_0(x) (\theta_t-\theta_0),  
\end{equation}
where the parameter $\theta_t$ is trained by 
\begin{equation}
    \theta_{t+1} = \theta_t -\eta G_0^{-1} \nabla_\theta \mathcal{L} (\theta_t). 
\end{equation}
The training dynamics of this linearized model is solvable and obtained by
\begin{equation}
     f_t^{lin}(x') = \bar{\Theta}_0(x',x)  \bar{\Theta}_0(x,x)^{-1}(I-(I-\eta \bar{\Theta}_0(x,x) )^t)(y-f_0(x))  +f_0(x'). \label{S24:0611}
\end{equation}
We evaluate the discrepancy between the original dynamics of wide neural networks $f_t$ and the above dynamics of linearized model $f_t^{lin}$. As is similar to the studies on GD \cite{jacot2018neural,lee2019wide}, we use   Gr\"onwall's inequality.
Precisely speaking, the previous works mainly focused on the continuous time limit and gave no explicit proof on the discrete time step. In the following, we show it by using a discrete analog of  Gr\"onwall's inequality.

\begin{thm}
\label{ThmA3}
Assume the same setting as in Theorem A.2. For $0<\eta\alpha<2$ and a sufficiently large $M$, with high probability, the discrepancy is given by
\begin{equation}
 \sup_t \|f_t^{lin}(x') - f_t(x')  \|_2 \lesssim A^3/\sqrt{M},
\end{equation}
on both training and test input samples $x'$.
\end{thm}

The notation $\lesssim$ hides the dependence on uninteresting constants.

{\it proof.}

\noindent 
{\bf (i) On training samples.} 

Let us denote the training error of the original model by $g_t(x) :=f_t(x)-y$.
and that of the linearized model by
$g_t^{lin}(x) :=f_t^{lin}(x)-y$.
Note that $f_t^{lin}-f_t = g_t^{lin}-g_t$. 
First, consider the trivial case of $\eta \alpha =1$.
By definition, we have $g_0^{lin}=g_0$ and $g_t^{lin}=0$ for $t>0$.  
we also have $||g_t||_2=(A'/\sqrt{M})^t$ ($t>0$) from Theorem A.2. Thus, we obtain the result. 

Next, consider the case of $\eta \alpha \neq 1$.
Denote a difference between time steps by $\Delta f_t := f_{t+1}-f_t$. We have 
\begin{align}
&\Delta (1-\eta \alpha)^{-t} (g^{lin}_t-g_t) \nonumber \\
&= \eta (1+\eta \alpha)^{-t-1}  [  (\alpha I-\tilde{J}_t G^{-1}_t J_t^\top/N)(g_t^{lin}-g_t)  -  (\alpha I-\tilde{J}_t G^{-1}_t J_t^\top/N  )g_t^{lin} ],  
\end{align}
where $\tilde{J}_t$ is the same as defined in (\ref{S14:0611}) and 
\begin{align}
    g_{t+1} = g_t + \tilde{J}_t (\theta_{t+1}-\theta_{t})  = ( I-\eta \tilde{J}_t G_t^{-1}J_t^\top/N) g_t.
\end{align}
We have also used  $g_{t+1}^{lin} = (1-\eta \alpha) g_t^{lin}$. 

 Taking the summation over time steps, we have
\begin{align}
g_{t+1}^{lin} -g_{t+1} 
&= \eta \sum_{s=0}^t (1-\eta \alpha)^{t-s}
[{ (\alpha I-\tilde{\Theta}_s)}(g_s^{lin}-g_s) - { (\alpha I-\tilde{\Theta}_s)} g_s^{lin}],    \label{S34:0608}
\end{align}
where we denote $\tilde{\Theta}_t := \tilde{J}_t G^{-1}_t J_t^\top/N$. Put $u_t:= \| g_t^{lin} -g_t \|_{{2} } $ and $Z_s:=\alpha I-\tilde{\Theta}_s$. By taking the norm of the above equation, we have 
\begin{equation}
   |1-\eta \alpha|^{-t} u_{t+1} \leq \eta \sum_{s=0}^t |1-\eta \alpha|^{-s} (\|Z_s \|_{{2} } u_s +\|Z_s \|_{{2} } \|g_s^{lin} \|_2 ).
    \label{S35:0608}
\end{equation}

We use the following discrete analogue of   Gr\"onwall's inequality (Theorem 4 in \cite{willett1965discrete}). Suppose  $\beta_t$, $\gamma_t$, and $U_{t+1}$ ($t=0,1,2,...$) are non-negative sequences of numbers with $\beta_0=\gamma_0=0$, and $c>0$. Then, the inequality 
\begin{equation}
    U_{t+1} \leq c + \sum_{s=0}^t \beta_s U_s + \gamma_t \label{S36:0608}
\end{equation}
implies that 
\begin{equation}
    U_{t+1} \leq (c+\gamma_t)  \prod_{s=0}^t (1+ \beta_s). \label{S37:0608}
\end{equation}
The inequality (\ref{S35:0608}) corresponds to (\ref{S36:0608}) by setting 
\begin{align}
U_t &=  |1-\eta \alpha|^{-t} u_t, \\    
\beta_s &= \eta \|Z_s \|_{{2} } \ \ \ \ (s>0), \\
\gamma_t&= \eta \sum_{s=0}^t |1-\eta \alpha|^{-s} \|Z_s \|_{{2} } \|g_s^{lin} \|_2 \ \ \ \ (t>0), \\
c&= \eta  \|Z_0\|_{{2} } \|g_0^{lin} \|_2. 
\end{align}
Note that we can set $\beta_0=0$ since we have $u_0=0$. The 
discrete analogue of Gr\"onwall's inequality  (\ref{S37:0608}) measures the discrepancy between the original and the linearized model. In the same way as in (\ref{S20:0929}), we have  
\begin{equation}
   \beta_s \leq A'  /\sqrt{M}. 
\end{equation}
Let us remind that we defined  $A'=4K A^2 R_0 /(1-|1-\eta \alpha|)$.
Similary, we have
\begin{align}
c &\leq \eta  \| \tilde{\Theta}_0 - \alpha I \|_{{2} } R_0 < R_0  A' /\sqrt{M}
\end{align}
and
\begin{align}
\gamma_t \leq \sum_{s=0}^t |1-\eta \alpha|^{-s} A'  \cdot    |1-\eta \alpha|^{s}R_0/\sqrt{M} 
= (t+1) R_0A' /\sqrt{M}.
\end{align}
Finally, the inequality (\ref{S37:0608}) gives \begin{align}
    u_{t+1} &\leq |1-\eta \alpha|^{t+1} (t+2)R_0A'/\sqrt{M} (1+A'/\sqrt{M})^t \\
    &=(t+2)R_0|1-\eta \alpha||1-\eta \alpha + (1-\eta \alpha)A'/\sqrt{M}|^t A' /\sqrt{M}. \label{S41:0611}
\end{align}
By taking a sufficiently large $M$, $|1-\eta \alpha + (1-\eta \alpha)A'/\sqrt{M}|^t$ converges to zero exponentially fast with respect to $t$.
Therefore, we have 
\begin{equation}
    \sup_t (t+2)|1-\eta \alpha + (1-\eta \alpha)A'/\sqrt{M}|^t = \mathcal{O}(1),
\end{equation}
where $\mathcal{O}(\cdot)$ is the big O notation. After all, we obtain $u_{t+1} \lesssim A^2/\sqrt{M}$. 

\noindent 
{\bf (ii) On test samples.} 

The discrepancy on the test samples $x'$ is upper bounded 
by the discrepancy on the training samples as follows. 
Note that we have 
\begin{equation}
    g_{t+1}(x')=g_t(x') - \bar{\eta} \tilde{J}_t(x')G_t^{-1}J_t^\top g_t 
\end{equation}
by using $\tilde{J}_t$, and 
\begin{equation}
    g_{t+1}^{lin}(x') = g_t^{lin}(x') -\bar{\eta} J_0(x')G_0^{-1}J_0^\top g_t^{lin}
\end{equation}
from Eq. (\ref{S24:0611}). Then, we have 
\begin{align}
&\|g_{t+1}^{lin}(x')-g_{t+1}(x') \|_2 \nonumber \\
&\leq \bar{\eta} \sum_{s=0}^t \| \tilde{J}_s(x')G_s^{-1}J_s^\top - J_0(x')G_0^{-1}J_0^\top \|_{{2} }
\|g^{lin}_s \|_2 +  \bar{\eta} \sum_{s=0}^t \| \tilde{J}_s(x')G_s^{-1}J_s^\top \|_{{2} }
\|g_s-g^{lin}_s \|_2 \\
&\leq \bar{\eta} R_0 \sum_{s=0}^t \| \tilde{J}_s(x')G_s^{-1}J_s^\top - J_0(x')G_0^{-1}J_0^\top \|_{{2} } |1-\eta \alpha|^{s} \\
&\ \  +  \bar{\eta} \sum_{s=0}^t( \| J_0(x')G_0^{-1}J_0^\top \|_{{2} } + \|\tilde{J}_s(x')G_s^{-1}J_s^\top - J_0(x')G_0^{-1}J_0^\top \|_{{2} }
)\|g_s-g^{lin}_s \|_2.\label{S47:0611} 
\end{align}
The Lipschitzness of Lemma A.1 and Condition 2 give 
\begin{align}
\|\tilde{J}_s(x')G_s^{-1}J_s^\top - J_0(x')G_0^{-1}J_0^\top \|_{{2} } \lesssim A^2/\sqrt{M}. \label{S48:0611} 
\end{align}
In addition, the inequality (\ref{S41:0611}) implies 
\begin{equation}
\| J_0(x')G_0^{-1}J_0^\top \|_{{2} } \sum_{s=0}^t \|g_s-g^{lin}_s \|_2 \lesssim A^3/\sqrt{M}.  \label{S49:0611} 
\end{equation}
Substituting (\ref{S48:0611}) and (\ref{S49:0611}) into (\ref{S47:0611}), we obtain $ \sup_t \|f_t^{lin}(x') - f_t(x')  \|_2 \lesssim A^3/\sqrt{M}$.
\qed

\subsection{Exact NGD}

As an example, we show that the exact (pseudo-inverse) FIM (8) satisfies Conditions 1 and 2. 
We have 
\begin{align}
    \bar{\Theta}(x',x) &= J(x') (J^\top J/N +\rho I)^{-1}  J^\top/N \\ 
 &= J(x')J^\top/N (JJ^\top/N +\rho I)^{-1}  \\ 
&= \Theta(x',x)(\Theta +\rho I)^{-1}. 
\end{align}
The NTK ($\Theta$) is positive definite \cite{jacot2018neural}. By setting $\rho=0$ and substituting the training samples to $x'$, 
we have Condition 1 with $\alpha=1$. 

Next, we show the exact FIM satisfies Condition 2. We neglect an uninteresting constant $1/N$ as long as it causes no confusion. 
We have 
\begin{align}
&\| G_0^{-1}J_0^\top -  G_s^{-1} J_s^\top \|_{{2} }  \nonumber \\ 
    &\leq \| J_0^\top (\Theta_0+\rho I)^{-1} -  J_s^\top (\Theta_s+\rho I)^{-1} \|_{{2} } \\
    &\leq   \| J_0-  J_s \|_{{2} }   \| (\Theta_0+\rho I)^{-1} \|_{{2} } +
     \|  J_s \|_{{2} }  \| (\Theta_0+\rho I)^{-1} - (\Theta_s+\rho I)^{-1} \|_{{2} }.  \label{S34:0529}
\end{align}
Here, we have
\begin{align}
\| (\Theta_0+\rho I)^{-1} - (\Theta_s+\rho I)^{-1} \|_{{2} }  \leq \| (\Theta_0+\rho I)^{-1} \|_{{2} } \| \Theta_0 - \Theta_s \|_{{2} }  \|  (\Theta_s+\rho I)^{-1} \|_{{2} }.
\label{S56:0609}
\end{align}
The NTK is positive definite \cite{jacot2018neural} and we have
\begin{equation}
    \| \Theta_0^{-1} \|_{{2} }  = 1/\lambda_{min}(\Theta_0),
\end{equation}
which may depend on the sample size, depth and hyper-parameters, but independent of widths.
Using the inequality $\sigma_{min}(A+B) \geq \sigma_{min}(A)-\sigma_{max}(B)$ where $\sigma$ denotes singular value, we obtain
\begin{equation}
\sigma_{min}(\Theta_s)  \geq \sigma_{min}(\Theta_0)-\|\Theta_s -\Theta_0\|_{{2} }.  \label{S58:0930}
\end{equation}
We have $\sigma_{min}(A)= \lambda_{min}(A)$ for a semi-positive definite matrix $A$. Note  that 
\begin{align}
\|\Theta_s -\Theta_0\|_{{2} }  \leq (\|J_s \|_{{2} } +\|J_0 \|_{{2} } )\|J_s-J_0 \|_{{2} } \leq 2K \|\theta_s -\theta_0 \|_2/\sqrt{M}.\label{S58:0610}
\end{align}
When  $\theta_s$ remain around the initialization with a finite radius, i.e., $\|\theta_s -\theta_0 \|\leq D$, we can take sufficiently small $\|\Theta_s -\Theta_0\|_{{2} }$ for a large $M$. 
Then, we obtain
\begin{equation}
\lambda_{min}(\Theta_s)  \geq \lambda_{min}(\Theta_0)/2 \label{S59:0610}
\end{equation}
from (\ref{S58:0930}). This means that $\Theta_s$ is positive definite and we can take $\rho=0$. 
The inequality (\ref{S56:0609}) becomes 
\begin{equation}
    \| \Theta_0^{-1} - \Theta_s^{-1} \|_{{2} } \leq \frac{4K}{\lambda_{min}(\Theta_0)^2}  \|\theta_s -\theta_0 \|_2/\sqrt{M}.
\end{equation}
Substituting this into (\ref{S34:0529}), we have 
\begin{align}
     \| J_0 \Theta_0^{-1} -  J_s \Theta_s^{-1} \|_{{2} } 
    &\lesssim   \| \theta_0-  \theta_s \|_{2}/\sqrt{M}. 
\end{align}
Thus,  the second inequality of Condition 2 holds.
From (\ref{S59:0610}), we also obtain the first inequality of Condition 2:
\begin{equation}
    \|G_s^{-1} J_s \|_{{2} } \leq \frac{2}{\lambda_{min}(\Theta_0)}K. \label{S60:0610}
\end{equation}

Since Conditions 1 and 2 hold, the NTK dynamics of exact NGD is given by Theorem A.3.

\section{Layer-wise NGD}

As preparation to prove Theorem 4.1, we define some notations and show lemmas.

We can represent the matrix $\Theta_l(x',x)(:= \nabla_{\theta_l} f_0(x') \nabla_{\theta_l} f_0(x)^\top/N)$ by a product between feedforward and backpropagated signals.
Note that the derivative $\nabla_\theta f$ is computed by the chain rule in a manner similar to the backpropagation algorithm: Given a single input $x$,  
\begin{align}
\frac{\partial f_{k}(x)}{\partial W_{l,ij}} &= \frac{\sigma_w}{\sqrt{M_l}} \cdot \delta_{l,i}^{(k)}(x) h_{l-1,j}(x), \ \ \frac{\partial f_{k}(x)}{\partial b_{l,i}} = \sigma_b \cdot \delta_{l,i}^{(k)}(x),  \\ 
\delta_{l,i}^{(k)}(x) &= \phi'(u_{l,i}(x)) \sum_{j} \delta_{l+1,j}^{(k)}(x) W_{l+1,ji},
\label{b_chain}
\end{align}
where $\delta_{l,i}^{(k)} := \partial f_{k}/\partial u_{l,i}$, and $f_k=u_{L,k}$ denotes the $k$-th unit of $u_L$ ($k=1,...,C$).  
We have $\delta_L^{(k)}=1$. We omit index $k$ of the output unit, i.e.,  $\delta_{l,i} = \delta_{l,i}^{(k)}$,  as long as the abbreviation causes no confusion. 

Now, we define two $N' \times N$ matrices as building blocks of $\Theta_l$ ($l=1,...,L-1$): 
\begin{equation}
    A_l(x',x) :=\frac{1}{M_l}h_l(x') h_l(x)^\top,
\end{equation}
 where  $h_l(x)$ represents an $N \times M_l$ matrix whose $i$-th row corresponds $i$-th input sample, and 
\begin{equation}
    B_l(x',x) :=\delta_l^{(k)}(x') \delta_l^{(k)}(x)^\top,
\end{equation}
 where  $\delta_l(x)$ represents an $N \times M_l$ matrix whose $i$-th row corresponds to $i$-th input sample. 
These two matrices have been investigated in the mean field theory of DNNs \cite{schoenholz2016,yang2020tensor}. In the infinite-width limit, we can analytically compute them as is overviewed in Section E. Note that the analytical kernel of $B_l$ is the same for any $k$.  
We also define $B_L:=1_{N'} 1_N^\top$ and $A_0:= X'X^\top/M_0$ where $X$ is a data matrix whose $i$-th row is the $i$-th sample vector $x$.
One can easily confirm
\begin{equation}
    \Theta_l(x',x)= I_C \otimes (\sigma_w^2 B_l(x',x) \odot  A_{l-1}(x',x)+ \sigma_b^2 B_l(x',x)).  \label{S67:0610}
\end{equation}
This kernel corresponds to the special case of NTK (\ref{S148:0609}) where only the $l$-th layer is used for training.

In our study, we need to investigate the positive definiteness of $\Theta_l$ to guarantee the convergence of layer-wise NGD. The following lemmas are helpful. 
\begin{lem}[\cite{jacot2018neural}]
Under Assumptions 1 and 2, $A_l$ ($l=1,...,L-1$) is positive definite in the infinite-width limit. 
\end{lem}
They proved this lemma in the following way. 
In the infinite-width limit, we have 
\begin{equation}
    A_l(x',x) =  \mathbb{E}_{u\sim \mathcal{N}(0,\sigma_w^2 A_{l-1}+\sigma_b^2 1 1^\top)} [\phi(u(x'))\phi(u(x))]. \label{S63:0609}
\end{equation}
The Gaussian integral over the inner product implies that when $\phi$ is non-constant and $A_{l-1}$ is positive definite, $A_l$ is positive definite. Therefore, the positive definiteness of $A_1$ leads to that of $A_l$ ($l=2,...,L-1$).
 When  $\phi$ is the non-polynomial Lipschitz function and $\|x \|_2=1$, we can prove the positive definiteness of $A_1$. 
Similarly, we obtain the following. 
\begin{lem}
Under Assumptions 1 and 2, $B_l$ ($l=1,...,L-1$) is positive definite in the infinite-width limit. 
\end{lem}
\noindent
Since $A_l$ is positive definite under Assumptions 1 and 2, the following matrix is also positive definite:
\begin{equation}
      \Xi_l(x',x) := \mathbb{E}_{u\sim \mathcal{N}(0,\sigma_w^2 A_{l-1}+\sigma_b^2 1 1^\top)} [\phi'(u(x'))\phi'(u(x))]. 
\end{equation}
The matrix $B_l(x',x)$ is given by $B_l=\sigma_w^2 \Xi_l \odot B_{l+1} $ in the infinite-width limit 
\cite{schoenholz2016,yang2020tensor}. Since the Hadamard product of two positive definite matrices is also positive definite,  $B_l$ is positive definite. 

Finally, we show the positive definiteness of $\Theta_l$ and an explicit formulation of $\bar{\Theta}$.  
\begin{lem}
 In the infinite-width limit on random initialization, (i) $\Theta_l$ is positive definite  for $l=2,...,L$, (ii) $\Theta_1$ is positive definite if $\sigma_b>0$ or if $A_0$ is full-rank, and (iii)  
when all of $\Theta_l$ are positive definite, the coefficient matrix of dynamics with $\rho=0$ is asymptotically equivalent to   
\begin{equation}
    \bar{\Theta}(x',x) =  \sum_{l=1}^L (\Sigma^{-1} 1_L)_l {\Theta}_l(x',x) {\Theta}_l^{-1}. \label{S70:0610}
\end{equation}
\end{lem}

\noindent
{\it Proof.}
Note that $\Theta_l$ is given by (\ref{S67:0610}), and that the Hadamard product between positive definite matrices is positive definite.  
For $l=2,...,L$,  $\Theta_l$ is positive definite because of Lemmas B.1. and B.2. For $l=1$, we need to pay attention to $A_0=X X^\top/M_0$ which may be singular.
if $\sigma_b>0$, $\Theta_1$ is positive definite because $B_1$ is positive definite. 
Thus, we obtain the results (i) and (ii).  

Now, we have
\begin{align}
    \bar{\Theta}(x',x) &= \frac{1}{N} J(x') (\frac{1}{N} S^\top(\Sigma \otimes I_{CN}) S +\rho I)^{-1} J^\top \label{S72:0609} \\ 
    &= \frac{1}{N} (1_L^\top \otimes I_{CN}) S(x')S^\top(\frac{1}{N} (\Sigma \otimes I_{CN})S S^\top +\rho I)^{-1}  (1_L \otimes I_{CN}) \label{S73:0609} \\
    &=  \sum_{l=1}^L (\Sigma^{-1} 1_L)_l {\Theta}_l(x',x) {\Theta}_l^{-1} \ \  \ \ \ \ (\rho=0). \label{S74:0609}
\end{align}
Note that $J^\top = S^\top (1_L \otimes I_{CN})$.
\qed

The condition of (ii) is not our interest but just a technical remark.  
We often use $\sigma_b>0$  in practice and
the condition holds. 
Even if $\sigma_b=0$ and $A_1$ is singular, the Hadamard product $\Theta_1$ can become positive definite depending on the training samples.

\subsection{Proof of Theorem 4.1}

By substituting the training samples to $x'$ in (\ref{S70:0610}), one can easily confirm that Condition 1 holds.

Next, we check Condition 2. 
We have 
\begin{align}
&\| G_0^{-1}J_0^\top -  G_s^{-1} J_s^\top \|_{{2} } \nonumber \\ 
    &\leq \| S_0^\top ((\Sigma \otimes I_{CN})S_0S_0^\top/N+\rho I)^{-1} -  S_s^\top ((\Sigma \otimes I_{CN})S_sS_s^\top/N+\rho I)^{-1} \|_{{2} } \|1_L \otimes I_{CN} \|_{{2} }  \\
    &\leq   \sqrt{L} (\| S_0-  S_s \|_{{2} }   \| (\Omega_0+\rho I)^{-1} \|_{{2} } +
     \|  S_s \|_{{2} }  \| (\Omega_0+\rho I)^{-1} -  (\Omega_s+\rho I )^{-1} \|_{{2} }),  \label{S77:0609}
\end{align}
where we denote $\Omega_s:=  (\Sigma \otimes I_{CN})S_sS_s^\top/N$. 
Here, we have
\begin{align}
&\|  (\Omega_0+\rho I)^{-1} -  (\Omega_s+\rho I)^{-1} \|_{{2} } \nonumber \\ &\leq \|  (\Omega_0+\rho I)^{-1} \|_{{2} } \| \Omega_0 - \Omega_s \|_{{2} }  \|  (\Omega_s+\rho I)^{-1} \|_{{2} } \\
&\leq \|   (\Omega_0+\rho I)^{-1} \|_{{2} } \max_l \|\Theta_l(s)-\Theta_l(0)  \|_{{2} } \| \Sigma  \|_{{2} }  \|   (\Omega_s+\rho I)^{-1} \|_{{2} }, \label{S77:0612}
\end{align}
where we denote $\Theta_l$ at time step $t$ by $\Theta_l(t)$. 
 Note that $\Theta_l(0)$ is positive definite from Lemma B.3, and that we supposed the positive definiteness of $\Sigma$. When $\rho=0$,  
\begin{equation}
    \| \Omega_0^{-1} \|_{{2} }  = (\min_l \lambda_{min}(\Theta_l(0)))^{-1} \lambda_{min}(\Sigma)^{-1}.
\end{equation}
Using the inequality $\sigma_{min}(A+B) \geq  \sigma_{min}(A)-\sigma_{max}(B)$ and $\sigma_{min}(\Omega_s)=\lambda_{min}(\Omega_s)$, we have 
\begin{align}
\lambda_{min}( \Omega_s)  &\geq \lambda_{min}( \Omega_0)-\| \Omega_s - \Omega_0\|_{{2} }  \\
&\geq \lambda_{min}( \Omega_0)- \max_l \| \Theta_l(s) - \Theta_l(0)\|_{{2} } \|\Sigma \|_{{2} }. \label{S81:0930}
\end{align}
In the same way as in (\ref{S58:0610}), we have
\begin{align}
\|\Theta_l(s) -\Theta_l(0)\|_{{2} }  &\leq (\|J_l(s) \|_{{2} } +\|J_l(0) \|_{{2} } )\|J_l(s)-J_l(0) \|_{{2} } \\ &\leq 2K \|\theta_s -\theta_0 \|_2/\sqrt{M}. \label{S82:0610}
\end{align}
Note that $J_l$  is the $l$-th block of $J$ and we can use Lemma A.1 because of $\| J_l\|_{{2} } \leq \| J\|_{F}$. 
In the same way as in (\ref{S59:0610}), we obtain
\begin{equation}
\lambda_{min}(\Theta_l(s))  \geq \lambda_{min}(\Theta_l(0))/2 \label{S84:0930}
\end{equation}
from (\ref{S81:0930}) and (\ref{S82:0610}).  
Then, we can set $\rho=0$ and the inequality (\ref{S77:0612}) becomes 
\begin{equation}
    \| \Omega_0^{-1} - \Omega_s^{-1} \|_{{2} } \lesssim  \|\theta_s -\theta_0 \|_2/\sqrt{M}.
\end{equation}
Substituting this into (\ref{S77:0609}), we obtain the second inequality of Condition 2: 
\begin{align}
     \| J_0 \Theta_0^{-1} -  J_s \Theta_s^{-1} \|_{{2} } 
    &\lesssim   \| \theta_0-  \theta_s \|_{2}/\sqrt{M}.
\end{align}
In addition, Ineq. (\ref{S84:0930}) implies the first inequality of Condition 2:
\begin{equation}
    \|G_s^{-1} J_s \|_{{2} } \leq  2 (\min_l \lambda_{min}(\Theta_l(0)))^{-1} \lambda_{min}(\Sigma)^{-1}\sqrt{L}K.
\end{equation}
We now finish the proof. 
\qed

\noindent 
{\bf Remark on the pseudo-inverse.} It may be helpful to remark that the deformation (\ref{S72:0609}-\ref{S74:0609}) corresponds to taking the pseudo-inverse of the layer-wise FIM. The similar deformation in the parameter space is given by 
\begin{align}
\Delta \theta &= G_t^{-1} J_t^\top (f-y) \\ 
&= S_t^\top (S_t S_t^\top)^{-1} ((\Sigma^{-1}1_L)\otimes I_{CN}) (f-y), \label{S88:0610}
\end{align}
where we have omitted an uninteresting constant $1/N$. Note that 
the Moore-Penrose pseudo-inverse of the layer-wise FIM ($\rho=0$) is  
\begin{equation}
    G_t^+ = S_t^\top (S_tS^\top_t)^{-1} (\Sigma \otimes I_{CN})^{-1} (S_tS^\top_t)^{-1} S_t.
\end{equation}
One can easily confirm that $G_t^+ \nabla_\theta \mathcal{L}$ is equivalent to the gradient (\ref{S88:0610}). 

\noindent 
{\bf Remark on singular $\Sigma$ of exact NGD.} 
 Theorem 4.1  assumed the positive definiteness of $\Sigma$. 
When $\Sigma$ is singular, $\Sigma$  inside the matrix inverse (\ref{S73:0609}) may cause instability as the damping term gets close to zero. This instability  was empirically confirmed in the singular tri-diagonal case. In contrast to Theorem 4,1, exact NGD (9) corresponds to $\Sigma=11^\top$ that is singular. It is noteworthy that this $\Sigma$ works as a special singular matrix in (\ref{S73:0609}). 
Since  $S_t^\top (\Sigma \otimes I_{CN}) S_t= J_t^\top J_t$, Eq. (S.72) becomes the pseudo-inverse of the exact NGD (9) as follows:
\begin{equation}
   (S_t^\top (\Sigma \otimes I_{CN}) S_t + \rho I )^{-1} J_t^\top =  J_t^\top ( J_t J_t^\top+ \rho I)^{-1}. \label{S90:0928}
\end{equation}
Thus, we can make $\Sigma$ inside of the inverse disappear and take the zero damping limit without any instability. Note that the transformation (\ref{S90:0928}) holds for any $J$. 
For general singular $\Sigma$, this instability seems essentially unavoidable. 
Potentially, there may exist a combination of a certain singular $\Sigma$ and a certain $J$ (e.g. certain network architecture) which can avoid the instability. Finding such an exceptional case may be an interesting topic, although it is out of the scope of the current work.

\subsection{Proof of Lemma 4.3}

Let us denote the $L\times L $ tri-diagonal matrix (19) by $\Sigma_L$. 
The Laplace expansion for determinants results in $|\Sigma_L| = |\Sigma_{L-1}|-|\Sigma_{L-2}|$ with $|\Sigma_3|=|\Sigma_4|=-1$. It is easy to confirm $|\Sigma_{3s+2}|=0$ while $|\Sigma_{3s}|=|\Sigma_{3s+1}| \neq 0$. 
As a side note, it is known that eigenvalues of $\Sigma_L$ are given by 
\begin{equation}
\lambda_\kappa= 1 + 2\cos \frac{\kappa \pi}{L+1},
\end{equation}
for $\kappa=1, ...,L$ \cite{noschese2013tridiagonal}.
Therefore, there is a zero eigenvalue when $\kappa \pi/(L+1)=2\pi/3$.
When $L=3s,3s+1$, all eigenvalues are non-zero. When $L=3s+2$,  we have $\lambda_{2(s+1)}=0$. 

Next, we compute $\alpha$ for $L=3s,3s+1$.  
In general, for a tri-diagonal Teoplitz matrix $\Sigma$ with the diagonal term of $a$ and the non-diagonal terms of $b$, we have [Corollary 4.4 \cite{da2001explicit}]
\begin{equation}
    1^\top \Sigma^{-1} 1 = \frac{L+2bs}{a+2b}, \ \  s := \frac{1+b(\sigma_1-\sigma_2)}{a+2b},
\end{equation}
where 
\begin{equation}
    \sigma_1 := \frac{1}{b} \frac{r_+^{L}-r_-^L }{r_+^{L+1}-r_-^{L+1}}, \ \  \sigma_2 := \frac{(-1)^{L+1}}{b} \frac{r_+-r_-}{r_+^{L+1}-r_-^{L+1}}, \ \ 
    r_\pm := \frac{a \pm \sqrt{a^2-4b^2}}{2b} . 
\end{equation}
\citet{da2001explicit} obtained this formula by using the explicit representation of $\Sigma^{-1}$ with the Chebyshev polynomials of the second kind. By substituting $a=b=1$, we have $r_\pm = \exp(i\pi/3)$ and we can easily confirm $\alpha =s$ for $3s$, and $\alpha=s+1$ for $3s+1$.

\subsection{K-FAC}

We suppose $C=1$ and $\sigma_b=0$ to focus on an essential argument of the NTK dynamics. 
 It is easy to generalize our results to  $\sigma_b>0$ as is remarked in Section B.3.3.

\subsubsection{Condition 1}

The block diagonal K-FAC (20) is defined  with 
\begin{align}
   A_l^* &:=\frac{\sigma_w^2}{NM_l} h_{l}^\top  h_{l}, \    B_l^* := \frac{1}{N}  \delta_{l}^\top  \delta_{l} \ \ (l<L),
\end{align}
where $h_l$ and $\delta_{l}$ denote $N \times M_l$ matrices whose $i$-th row corresponds to the $i$-th input sample. We set $B_L^*=1/N$. 
Then, the $st$-th entry of $\bar{\Theta}(x',x)$ is given by
\begin{align}
\bar{\Theta}(x',x)_{st}&=  \sum_l \frac{\sigma_w^2}{N M_{l-1}} \delta_{l}(x'_s)^\top (B_{l}^* +\rho I )^{-1}\delta_{l}(x_t) h_{l-1}(x'_s)^\top  (A_{l-1}^* +\rho I )^{-1} h_{l-1}(x_t). \label{S96:0610}
\end{align}

Let us represent the derivative by 
\begin{equation}
 \nabla_{\theta_l} f (x_n) = \frac{\sigma_w}{\sqrt{M_{l-1}}} (\delta_{l}^\top e_n) \otimes  (h_{l}^\top e_n), \label{S95:0610}  
\end{equation}
where $e_n$ is a unit vector whose $n$-th entry is 1 and otherwise 0.
We have
 \begin{align}
\delta_{l}(x'_s)^\top (B_{l}^* +\rho I )^{-1}\delta_{l}(x_t) &= (\delta_l(x')^\top e_s)^\top (\delta_{l}^\top \delta_l/N +\rho I )^{-1} \delta_l ^\top e_t\\
&= e_s^\top B_l(x',x)(B_l/N+\rho I)^{-1} e_t \\
& = N  (B_l(x',x)B_l^{-1})_{st}  \ \ (\rho=0),
\end{align}
for $l\geq 1$. In the last line, we use the positive definiteness shown in Lemma B.2.
Similarly, for $l\geq 2$, 
 \begin{align}
 \frac{\sigma_w^2}{M_l}h_{l}(x'_s)^\top  (A_{l}^* +\rho I )^{-1} h_{l}(x_t)  &=    \frac{\sigma_w^2}{M_l} (h_{l}(x')^\top e_s)^\top (\sigma_w^2 h_{l}^\top h_{l}/(M_lN) +\rho I )^{-1} (h_{l}^\top e_t)\\
&= \sigma_w^2 e_s^\top A_l(x',x)( \sigma_w^2 A_l/N+\rho I)^{-1} e_t  \\
& =N (A_l(x',x)A_l^{-1})_{st}  \ \ (\rho=0), \label{S101:0610}
\end{align}
 where we use the positive definiteness shown in Lemma B.1. $A_0$ depends on settings of input data as follows. 

\noindent
{\bf (i) Case of $ M_0 \geq N$}

Assume that the input samples are linearly independent (that is, full-rank $A_0$).  
Then, we can take $\rho=0$ and we obtain (\ref{S101:0610})  for $l=1$ as 
\begin{equation}
     N  \cdot X' X^\top ( X X^\top)^{-1}.  \label{S102:0610}
\end{equation}
After all, 
we have 
\begin{equation}
  \bar{\Theta}(x',x) =  N \sum_{l=1}^L   \mathcal{B}_l \odot \mathcal{A}_{l-1}, \label{S103:0610}
\end{equation}
where 
\begin{equation}
 \mathcal{B}_l:=B_l(x',x)B_l^{-1}, \ \  \mathcal{A}_l:=A_l(x',x)A_l^{-1},
\end{equation}
for $0<l<L$ and $\mathcal{B}_L:=1_{N'} 1_N^\top$.
By setting the training samples to $x'$, we have    
\begin{equation}
    \bar{\Theta} = \alpha I, \ \alpha =NL. 
\end{equation}

\noindent {\bf (ii) Case of $ M_0 < N$}

While we can take the pseudo-inverse of $X$ in (\ref{S102:0610}) for $M_0 \geq N$, 
$XX^\top$ becomes singular for $ M_0 < N$ and we need to use $A^*_0$ in the K-FAC gradient. 
Assume that $A^*_0$ is full-rank. 
By setting $\rho=0$, $\bar{\Theta}(x',x)$ becomes (\ref{S103:0610}) with 
\begin{equation}
\mathcal{A}_0(x',x)=X' (X^\top   X)^{-1} X^\top. 
\end{equation}
Therefore, for the training samples, we obtain
\begin{equation}
 \frac{1}{N}\bar{\Theta} = (L-1)I + (I \odot X (X^\top   X)^{-1} X^\top).
\end{equation}
This means that the isotropic condition does not hold in naive settings. 
 \citet{zhang2019fast} pointed out a similar phenomenon in K-FAC training of the first layer of a shallow ReLU network. 
 Fortunately, they found that by using pre-processing of $X$ known as the Forster transformation, we can transform $X$ into $\bar{X}$ such that $\bar{X}^\top \bar{X}  =\frac{N}{M_0} I$ while keeping the normalization of each sample ($\|\bar{x} \|_2=1$; Assumption 2). After the Forster transformation, we have  
 \begin{equation}
    \mathcal{A}_0(x',x)= \frac{M_0}{N} X'\bar{X}^\top   
 \end{equation}
and the isotropic condition as
 \begin{equation}
\bar{\Theta} = \alpha I, \ \alpha = N(L-1) + M_0. 
 \end{equation}

\subsubsection{Condition 2}

Next, we check Condition 2. By using the representation (\ref{S95:0610}),
the $l$-th layer part of $G^{-1}J^\top$ is given by 
\begin{align}
 &(B^*_l+\rho I)^{-1} \otimes (A^*_{l-1}+\rho I)^{-1}
    (\nabla_{\theta_l} f e_n) \nonumber \\
    &=     \underbrace{\left(\left(\delta_l^\top (B_l/N+\rho I)^{-1}\right)\otimes \left(\frac{\sigma_w}{\sqrt{M_{l-1}}} h_{l-1}^\top (\sigma_w^2A_{l-1}/N+\rho I)^{-1}\right)\right)}_{=:Z_l} (e_n \otimes e_n). 
\end{align} 
Therefore, 
\begin{align}
\| G_0^{-1}J_0^\top -  G_s^{-1} J_s^\top \|_{{2} } \nonumber 
    &\leq \max_l  \| Z_l(0) \Lambda -  Z_l(s) \Lambda \|_{{2} } \|1_L \otimes I_{CN} \|_{{2} } \\
    &\leq \max_l   \| Z_l(0) -  Z_l(s) \|_{{2} }  \sqrt{NL}, \label{S112:0610}
\end{align}
where $\Lambda$ is an $N^2 \times N$ matrix whose $i$-th column is $e_i\otimes e_i$. 
Define
\begin{align}
  Z^B(s)&=\delta_l(s)^\top (B_l(s)/N+\rho I)^{-1}, \\ 
Z^A(s) &=  \frac{\sigma_w}{\sqrt{M_{l-1}}} h_{l-1}(s)^\top (\sigma_w^2A_{l-1}(s)/N+\rho I)^{-1}.
\end{align}
As we discussed in the above subsection, $Z^A(s)$ at $l=1$ is given by $(X X^\top)^{-1}X$ for $M_0 \geq N$ and  $X (X^\top X)^{-1}$  for $M_0 < N$. 
We have 
\begin{align}
  &\| Z_l(0) -  Z_l(s) \|_{{2} } \nonumber \\
  &\leq 
 \|Z^B(s) \otimes Z^A(s) - Z^B(0) \otimes Z^A(0)  \|_{{2} }  \\ 
 &\leq \|Z^B(s)  - Z^B(0)\|_{{2} }  \| Z^A(s)  \|_{{2} } + \| Z^B(0)  \|_{{2} } \|Z^A(s)  - Z^A(0)\|_{{2} }. \label{S115:0610}
\end{align}
Here, we can use the Lipschitzness in the same way as in (\ref{S34:0529}). For example, we have 
\begin{align}
\|Z^B(s)  - Z^B(0)\|_{{2} } 
    &\leq  \| \delta_l(0)-  \delta_l(s) \|_{{2} }   \| (B_l(0)/N+\rho I)^{-1} \|_{{2} } \nonumber \\ 
    &\ \ \ \ \ \ \ \ \ \    +
     \|  \delta_l(s) \|_{{2} }  \| (B_l(0)/N+\rho I)^{-1} - (B_l(s)/N+\rho I)^{-1} \|_{{2} }).
\end{align}
Lemma A.1 gives Lipschitz bounds of terms including  $\delta_l$. 
From Lemma B.2, we have
\begin{equation}
    \| B_l(0)^{-1} \|_{{2} }  = 1/\lambda_{min}(B_l(0)).
\end{equation}
By the same calculation as in (\ref{S56:0609}), we  have 
\begin{equation}
    \| B_l(0)^{-1} - B_l(s)^{-1} \|_{{2} } \lesssim  \|\theta_s -\theta_0 \|_2/\sqrt{M}.
\end{equation}
In this way, we can obtain the Lipschitz bound of $\|Z^B(s)  - Z^B(0)\|_{{2} }$. Similarly, we obtain the bounds of  $\|Z^A(s)  - Z^A(0)\|_{{2} }$, $\|Z^A(s)\|_{{2} }  \|$ and $\|Z^B(s)\|_{{2} }$. 
They give a bound of (\ref{S112:0610}) via (\ref{S115:0610}), and we obtain the second inequality of Condition 2: 
\begin{equation}
\| G_0^{-1}J_0^\top -  G_s^{-1} J_s^\top \|_{{2} } \lesssim  \|\theta_s -\theta_0 \|_2/\sqrt{M}.
\end{equation}
 In the same argument, we also obtain the first inequality of Condition 2 via
\begin{equation}
\|G_s^{-1} J_s \|_{{2} } \leq \max_l \|Z_l(s)\|_{{2} } \sqrt{NL}.
\end{equation}
After all, we confirm both Conditions 1 and 2 are satisfied, and the NTK dynamics is given by $f_t^{lin}(x')$ in Theorem A.3

\subsubsection{K-FAC with bias terms}
We can obtain the K-FAC with bias terms by replacing the vector $ \frac{\sigma_w}{\sqrt{M_l}}h_l(x) \in \mathbb{R}^{M_l}$ with $[\frac{\sigma_w}{\sqrt{M_l}} h_l(x);\sigma_b] \in  \mathbb{R}^{M_l+1}$. 
For $M_0 \geq N$,  we just need to replace $\sigma_w^2 A_l(x',x)$ by $\sigma_w^2 A_l(x',x) + \sigma_b^2 11^\top$ for all $l \geq 0$.  
 This approach is applicable to $M_0 < N$ as well. 
 We can regard $[\frac{\sigma_w}{\sqrt{M_0}} x_n;\sigma_b]$ as  new input samples and apply the Forster transformation to them. 
 However, it may be unusual to normalize $x_n$ with such an additional one dimension ($\sigma_b$). 
One alternative approach is to use the following block FIM; 
\begin{equation}
    G= \begin{bmatrix}  G_{\text{K-FAC}} & 0 \\ 0 & \nabla_b f \nabla_b f^\top/N^2 + \rho I \end{bmatrix},
    \end{equation}
where the weight part is given by K-FAC  and the bias part is given by a usual FIM.    
In this case, since the weight part does not include the additional dimension, we can use the Forster transformation as usual. 
We have
\begin{equation}
\frac{1}{N} \bar{\Theta}(x',x)=\sum_{l=1}^L   \mathcal{B}_l \odot (\mathcal{A}_{l-1}+11^\top).
\end{equation}

\subsection{Min-norm solution}
Let us denote $E_\lambda(\theta):= \frac{1}{2N} \|y-  J_0\theta\|^2_2+ \frac{\lambda}{2} \theta^\top G_0 \theta$. For $\lambda>0$, 
it has a unique solution  $\theta^*_\lambda := \mathrm{argmin}_\theta E_{\lambda>0}(\theta)$. 
 After a straight-forward linear algebra, $\nabla_\theta E_{\lambda>0}(\theta)=0$ results in  
 \begin{align}
   \theta^*_\lambda &=  (\lambda G_0 +J_0^\top J_0/N)^{-1} J_0^\top y/N \\
   &= G_0^{-1} J_0^\top (\lambda I + J_0 G_0^{-1} J_0^\top /N)^{-1}y/N \\ 
   &= \frac{1}{\lambda + \alpha} G_0^{-1} J_0^\top y/N,
 \end{align}
 where we used a  matrix formula $(A+BB^\top)^{-1} B =A^{-1}B(I+B^\top A^{-1}B)^{-1}$ (Eq.(162) in \cite{cookbook}) and the isotropic condition $J_0 G_0^{-1} J_0^\top/N = \alpha I$.  
After all,  $\lim_{\lambda \to 0} \theta^*_\lambda$ is equivalent to the NGD solution $\theta_\infty$.

\section{Unit-wise NGD}

First, we show that the unit-wise FIM satisfies Condition 1 under a specific assumption.  Second, we reveal that Condition 2 holds with keeping a finite damping term $\rho>0$.  
 Finally, by taking the zero damping limit and using Theorem A.3, we prove the fast convergence of unit-wise NGD (Theorem 5.1). 

We suppose $C=1$. We also assume $M_0 \geq N$, and linear independence of input samples (that is,  full-rank $A_0$). 
 The case of $M_0 < N$ is discussed in Section C.2.2.

\subsection{Condition 1}

We show that under the following assumption, Condition 1 holds: 
\begin{as}[the gradient independence assumption \cite{schoenholz2016,yang2017,xiao2018dynamical,karakida2019normalization,karakida2018universal}]
{\it When one evaluates a summation over $\delta_{l,i}(x_n)$ ($i=1,...,M_l$),  one can replace 
weight matrices $W_{l+1,ji}$ in the chain rule (\ref{b_chain}) with a fresh i.i.d. copy, i.e.,  $\tilde{W}_{l,ji}  \overset{\text{i.i.d.}}{\sim}\mathcal{N}(0,1)$.}
\end{as}
Assumption C.1 has been used as an essential technique of the mean field theory for DNNs.
This assumption makes random variables $\delta_{l,i}$ ($i=1,...,M_l$) independent with each other, and enables us to use the law of large numbers or the central limit theorem in the infinite-width limit.
 \citet{schoenholz2016} found that some order parameters (e.g., $\sum_i \delta_{l,i}(x_n)^2$) obtained under this assumption show a very good agreement with experimental results. Excellent agreements between  the theory and experiments have been also confirmed in various architectures \cite{yang2017,xiao2018dynamical} and algorithms \cite{karakida2019normalization}.  
Thus,  Assumption C.1 will be useful as the first step of the analysis.

\begin{lem}
Suppose Assumption C.1.
on random initialization, for a sufficiently large $M$ and constants $\gamma_l>0$, the unit-wise FIM satisfies
\begin{equation}
\bar{\Theta} = \alpha I, \ \ \alpha=\sum_{l=1}^{L-1} \gamma_l M_l, \label{S119:0611}
\end{equation}
in the zero damping limit ($\rho \rightarrow 0$).
\end{lem}

\noindent 
{\it Proof.} We can represent the unit-wise FIM (23) by using 
\begin{equation}
  S_{\text{unit},t}:=  \begin{bmatrix}
 D_1           &     &         &  O \\
   &  D_2  &      &          \\
   &  & \ddots &   \\
 O  &        &       &  D_L       
 \end{bmatrix}, \ \ 
  D_l:=  \begin{bmatrix}
 \nabla_{\theta_1^{(l)}} f_t          &     &         &  O \\
   &  \nabla_{\theta_2^{(l)}} f_t  &      &          \\
   &  & \ddots &   \\
 O  &        &       &   \nabla_{\theta_{M_l}^{(l)}} f_t
 \end{bmatrix}.
 \end{equation}
 In this proof, we consider the random initialization and omit the index of $t=0$. 
$D_l$ is an $M_{l} N \times M_l(M_{l-1}+1)$ block matrix whose diagonal blocks are given by $\nabla_{\theta_i^{(l)}} f $, an $N \times (M_{l-1}+1)$ matrix. 
Note that $J^\top = S^\top_{\text{unit}} (1_{M'} \otimes I_{N})$ with $M':=\sum_{l=1}^L M_l$.
We have
\begin{equation}
 \bar{\Theta}=\sum_{l=1}^{L}\sum_{i=1}^{M_l}\Theta_{l,i}(\Theta_{l,i}+\rho I)^{-1},
\end{equation}
where we define $\Theta_{l,i}:= \nabla_{\theta_i^{(l)}} f \nabla_{\theta_i^{(l)}} f^\top/N$ ($N \times N$ matrix). 
Here, we need to be careful on the positive definiteness of $\Theta_{l,i}$. We have 
\begin{equation}
    \Theta_{l,i}= \mathrm{diag}(\delta_{l,i})A_{l-1} \mathrm{diag}(\delta_{l,i}),
\end{equation}
where $\mathrm{diag}(y)$ denotes a diagonal matrix with diagonal entries given by entries of the vector $y$. 
If any entry of $\delta_{l,i}$ takes zero, $\Theta_{l,i}$ is singular.  
For instance, in ReLU networks, we will be likely to get $\delta_{l,i} (x_n) = 0$  because $\phi'(u)=0$ for $u\leq 0$.

When $\delta_{l,i}(x_n) \neq 0$ for $n=n_1, n_2, ..., n_r$, we rearrange $\delta_{l,i}$ ($N$ dimensional vector) into another $N$ dimensional vector $\bar{\delta}_{l,i}$ whose first $r$ entries take non-zero and the others take zero. 
Because this is just a rearrangement of the entry, we can represent it by $\delta_{l,i} = Q \bar{\delta}_{l,i}$ where $Q$ is a certain regular matrix given by a product of elementary permutation matrices for entry switching transformations. Then, we have $\mathrm{diag}(\delta_{l,i}) = Q \mathrm{diag}(\bar{\delta}_{l,i}) Q$. 
Note that, because the inverse of the elementary permutation matrix is itself, we have $Q=Q^{-1}=Q^\top$.

Using this rearrangement notation of the entries, we have 
\begin{equation}
   \Theta_{l,i} =  Q \mathrm{diag}(\bar{\delta}_{l,i}) \bar{A}_{l-1} \mathrm{diag}(\bar{\delta}_{l,i}) Q,
\end{equation}
with $\bar{A}_{l-1} := Q A_{l-1} Q$. 
We can represent it by  
\begin{equation}
   \mathrm{diag}(\bar{\delta}_{l,i}) \bar{A}_{l-1} \mathrm{diag}(\bar{\delta}_{l,i}) =\begin{bmatrix}  \mathrm{diag}(\bar{\delta}'_{l,i}) \bar{A}_{l-1}' \mathrm{diag}(\bar{\delta}'_{l,i}) & O \\ O & O    \end{bmatrix}, \label{S123:0611}
\end{equation}
where $\bar{\delta}'_{l,i} \in \mathbb{R}^{r}$ and $\bar{A}_{l-1}' \in \mathbb{R}^{r\times r}$ denote the non-zero part. 
Then, we have 
\begin{align}
\Theta_{l,i}(\Theta_{l,i}+\rho I)^{-1}  &=   Q\begin{bmatrix}  \mathrm{diag}(\bar{\delta}'_{l,i}) \bar{A}_{l-1}' \mathrm{diag}(\bar{\delta}'_{l,i}) (\mathrm{diag}(\bar{\delta}'_{l,i}) \bar{A}_{l-1}' \mathrm{diag}(\bar{\delta}'_{l,i})+\rho I)^{-1} & O \\ O & O    \end{bmatrix}Q,
\end{align}
where we use $O \cdot (I/\rho) =O$ for $\rho>0$ for 
the zero part of (\ref{S123:0611}). This means that the one-sided limit is given by 
\begin{equation}
\lim_{\rho \rightarrow 0^+}  \Theta_{l,i}(\Theta_{l,i}+\rho I)^{-1} =  Q\begin{bmatrix} I_r & O \\ O & O    \end{bmatrix}Q.
\end{equation}
We have used that $\bar{A}'_l$, i.e., a submatrix of $\bar{A}_l$,  is positive definite because the original matrix $A_l$ is positive definite by Lemma B.1.
Since we can rearrange the matrix into the original alignment with the operation $Q (\cdot)Q$, we have 
\begin{equation}
Q\begin{bmatrix} I_r & O \\ O & O    \end{bmatrix}Q =  \mathrm{diag} (1_{\delta_{l,i} \neq 0}(\delta_{l,i})),
\end{equation}
where we define an indicator function by $ 1_{A}(x):= 1$ (when $A$ holds), $0$ (otherwise).  

After all, we have
\begin{equation}
 \lim_{\rho \rightarrow 0^+} \bar{\Theta}=\sum_{l=1}^{L} \sum_{i=1}^{M_l}  \mathrm{diag} (1_{\delta_{l,i} \neq 0}(\delta_{l,i})). \label{S132:0612}
\end{equation}
Note that we have $M_L=1$ and the contribution of the $L$-th layer in (\ref{S132:0612}) is negligible at a large $M$. We have $\delta_{l,i} = \phi'(u_{l,i}) \sum_{j} \delta_{l+1,j} \tilde{W}_{l+1,ji}$.
Since ${W}_l$ is a Gaussian random matrix,  
$u_{l,i}$ is Gaussian random variable (for $i=1,...,M_l$) \cite{schoenholz2016,lee2019wide}. As is used in these previous works, its variance ($q_l:= \sum_{i=1}^{M_l} u_{l,i}^2/M_l$) is given by 
\begin{equation}
    q_{l+1} = \frac{\sigma_w^2}{\sqrt{2 \pi q_l}} \int du \phi(u)^2 \exp  \left(- \frac{u^2}{2q_l}\right)  + \sigma_b^2,
\end{equation}
with $q_0 = \|x_n \|^2/M_0=1/M_0$. 
When we evaluate the summation over $\delta_{l,i}$ in (\ref{S132:0612}), the indicator function requires a careful evaluation on the case of $\delta_{l,i}=0$. Let us denote $\tau_{l,i}:= \sum_{j} \delta_{l,j} \tilde{W}_{l,ji}$. We have $\delta_{l,i} = \phi'(u_{l,i}) \tau_{l+1,i}$. 
Here, we use Assumption C.1 to decouple the contribution of $\phi'(u_{l,i})$ and that of $\tau_{l+1,i}$.  We have 
\begin{equation}
   \tau_{l+1,i} \sim  \mathcal{N} (0, \  \sum_j \delta_{l+1,j}(x_n)^2), \label{S128:0611} 
\end{equation}
for $i=1,...,M_l$. In the large $M$ limit, $\sum_j \delta_{l+1,j}(x_n)^2$ converges to a constant known as 
the order parameter \cite{schoenholz2016,yang2020tensor}. Because Assumption C.1 enables us to take the Gaussian integral over $u_{l,i}$ and $\tau_{l+1,i}$ independently, we obtain
\begin{align}
    \frac{1}{M_l}\sum_{i=1}^{M_l} 1_{\delta_{l,i}(x_n) \neq 0}(\delta_{l,i}(x_n)) &= \frac{1}{\sqrt{2\pi q_l}}  \int du 1_{\phi'(u) \neq 0}(u) \exp \left(- \frac{u^2}{2q_l}\right) \\
    &=: \gamma_l. \label{S131:0611}
\end{align}
Since this holds independently of the sample index $n$, we obtain (\ref{S119:0611}). \qed

From this Lemma, one can see that Condition 1 holds. 
The constants $\gamma_l$ depend on the shape of the activation function. For instance, when one uses
activation  functions with $\phi'(x)^2 \neq 0$ for almost everywhere (e.g. Tanh), we have $\gamma_l=1$. 
In Section C.3.3, we explicitly show $\gamma_l$ in the case of (shifted-) ReLU.  
Figure S.2 shows an excellent agreement with the numerical values of $\alpha$ and our analytical solutions obtained by (\ref{S131:0611}).

\noindent
 {\bf Remark on the justification of Assumption C.1:}
    After the submission of our paper, \citet{yang2020tensor} rigorously justified that various calculations based on the gradient independence assumption results in correct answers. In particular, Theorem 7.2 \cite{yang2020tensor} justifies our evaluation of (\ref{S131:0611}) when the activation function is polynomially bounded.  The replacement with the fresh i.i.d. copy naturally appears through a Gaussian conditioning technique even in the exact calculation without the gradient independence assumption. It leads to the same Gaussian integrals and decoupling between $u_{l,i}$ and $\tau_{l+1,i}$ as in (\ref{S132:0612})-(\ref{S131:0611}).

\subsection{Condition 2 and Proof of Theorem 5.1}

\subsubsection{Condition 2}

\begin{lem} 
 There is a constant $A>0$ such that for a sufficiently large $M$, a damping term $\rho>0$ and every $D>0$, the following holds with high probability,
\begin{equation}
 \left\{\begin{array}{ll}
    \bar{\eta}\| G_{{\text{\rm  unit}},s}^{-1} J_s^\top \|_{\text{\rm  op}}  &\leq A \rho^{-1}  \\
    \bar{\eta}\| G_{{\text{\rm  unit}},0}^{-1}J_0^\top -G_{{\text{\rm  unit}},s}^{-1} J_s^\top \|_{\text{\rm op}}  &\leq A\rho^{-2}\|\theta_s-\theta_0\|_2 /\sqrt{M}
\end{array} \quad \quad \forall \theta_s \in B\left(\theta_{0}, D \right),\right. \label{S137:0612}
\end{equation}
where the learning rate is $\eta=c/M$ for $c>0$.
\end{lem}

\noindent 
{\it Proof. } 
For $\eta =c/M$, we have 
\begin{align}
    &\bar{\eta} \| G_{0}^{-1}J_0 - G_{s}^{-1} J_s\|_{{2} } \nonumber \\
    &\leq \bar{\eta} \|S(0)^\top (  S(0) S(0)^\top/N + \rho I)^{-1}     -  S(s)^\top (  S(s) S(s)^\top/N + \rho I)^{-1}  \|_{{2} } \| 1_{M'} \otimes I_{CN}\|_{{2} }  \\
    &\leq  c' \max_{l,i} ( \| J_{l,i}(0)-  J_{l,i}(s) \|_{{2} }   \| (\Theta_{l,i}(0)+\rho I)^{-1} \|_{{2} } \nonumber \\ 
    & \ \ \ \ \ \ \ \ \ \  \ \ \ \ \ \ \ \ \ \ \ \  + \|  J_{l,i}(s) \|_{{2} }  \| (\Theta_{l,i}(0)+\rho I)^{-1} -(\Theta_{l,i}(s)+\rho I)^{-1} \|_{{2} } ), \label{S139:0612}
\end{align}
where we denote $S_{\text{unit},s}$ by $S(s)$, the Jacobian $\nabla_{\theta_i^{(l)}} f_s$ by $J_{l,i}(s)$, and an uninteresting constant by $c'$.  Here, we have 
\begin{align}
&\| (\Theta_{l,i}(0)+\rho I)^{-1} - (\Theta_{l,i}(s)+\rho I)^{-1} \|_{{2} }  \nonumber \\
&\leq \| (\Theta_{l,i}(0)+\rho I)^{-1} \|_{{2} } \| \Theta_{l,i}(0) - \Theta_{l,i}(s) \|_{{2} }  \|  (\Theta_{l,i}(s)+\rho I)^{-1} \|_{{2} }. \label{S136:0611}
\end{align}
Using the inequality $\|(A+B)^{-1}\|_{{2} } \leq 1/(\lambda_{min}(A)+\lambda_{min}(B)) \leq 1/\lambda_{min}(B)$, we obtain
\begin{equation}
\| (\Theta_{l,i}(0)+\rho I)^{-1} \|_{{2} }  \leq 1/\rho. \label{S144:0612}
\end{equation}
Using the inequality $\sigma_{min}(A+B) \geq  \sigma_{min}(A)-\sigma_{max}(B)$, we obtain 
\begin{equation}
\lambda_{min}(\Theta_{l,i}(s)+\rho I)  \geq \lambda_{min}(\Theta_{l,i}(0)+\rho I)-\|\Theta_{l,i}(s) -\Theta_{l,i}(0)\|_{{2} }. \label{S148:0930}
\end{equation}

In the same way as in (\ref{S58:0610}), we have
\begin{align}
\|\Theta_{l,i}(s) -\Theta_{l,i}(0)\|_{{2} }  &\leq (\|J_{l,i}(s) \|_{{2} } +\|J_{l,i}(0) \|_{{2} } )\|J_{l,i}(s)-J_{l,i}(0) \|_{{2} } \\
&\leq 2K \|\theta_s -\theta_0 \|_2/\sqrt{M}. \label{S146:0612}
\end{align}
Note that $J_{l,i}$  is a block of $J$. We have $\| J_{l,i}\|_{{2} } \leq \| J\|_{F}$ and can use Lemma A.1. 
In the same way as in (\ref{S59:0610}), we obtain
\begin{equation}
\lambda_{min}(\Theta_{l,i}(s)+\rho I)    \geq \lambda_{min}(\Theta_{l,i}(0)+\rho I)  /2 \geq \rho/2 \label{S147:0612}
\end{equation}
from (\ref{S148:0930}) and (\ref{S146:0612}). 
Substituting (\ref{S144:0612}), (\ref{S146:0612}) and (\ref{S147:0612}) into the inequality (\ref{S136:0611}), we have
\begin{equation}
  \| (\Theta_{l,i}(0)+\rho I)^{-1} - (\Theta_{l,i}(s)+\rho I)^{-1} \|_{{2} } \leq 2K \|\theta_s -\theta_0 \|_2\rho^{-2}/\sqrt{M}. 
\end{equation}
Substituting this into (\ref{S139:0612}), we obtain the second inequality of Condition 2:
\begin{align}
    \bar{\eta} \| G_{0}^{-1}J_0 - G_{s}^{-1} J_s\|_{{2} } 
    &\leq A \rho^{-2}  \| \theta_0-  \theta_s \|_{2}/\sqrt{M}.
\end{align}
In addition, Ineq. (\ref{S147:0612}) implies the first inequality of Condition 2:
\begin{equation}
   \eta \|G_s^{-1} J_s \|_{{2} } \leq A \rho^{-1},
\end{equation}
where an uninteresting constant $A$ is independent of $M$ and $\rho$. 
We obtain the desired result. \qed

\subsubsection{Convergence of training dynamics (Proof of Theorem 5.1)}

Let us consider a zero damping limit of $\rho=1/M^\varepsilon$ ($\varepsilon>0$). Under the zero damping limit, Lemma C.2 holds and the isotropic condition is satisfied.  
 Regarding Condition 2,  note that we keeps $\rho>0$ in Lemma C.3 while we exactly set $\rho=0$ in Condition 2 of other FIMs. The effect of $\rho>0$ on the bound appears as $A\rho^{-1}$ and $A\rho^{-2}$ in  Lemma C.3. 
 When $\rho$ is small,  we have $\rho^{-1}<\rho^{-2}$ and  the first inequality of (\ref{S137:0612}) is also bounded by $A\rho^{-2}$. 
 Therefore,  $A$ in Theorem A.3 is replaced by $A\rho^{-2}$ in unit-wise NGD. Note that in Theorem A.3 and its proof, $A$ appears in the form of $A^2/\sqrt{M}$, or $A^3/\sqrt{M}$ at the worst case. By taking the zero damping limit with $0<\varepsilon<1/12$, we obtain the bound of Theorem A.3 as follows:  
\begin{equation}
 \sup_t \|f_t^{lin} - f_t  \|_2 \lesssim A^3\rho^{-6}/\sqrt{M} = A^3/M^{1/2(1-12\varepsilon)}.  
\end{equation}
 After all, the training dynamics is given by $f_t^{lin}$ in the infinite-width limit. \qed

We have also confirmed that the training dynamics obtained in Theorem 5.1 show an excellent agreement with numerical experiments of training. See Figure S.1 in Section C.3.2.  

\noindent
{\bf Remark.}  First, note that the coefficient matrix on test samples $x'$ becomes   
\begin{equation}
 \bar{\Theta}(x',x)=\sum_{l=1}^{L}\sum_{i=1}^{M_l}
  \mathrm{diag}(\delta_{l,i}(x'))A_{l-1}(x',x) \mathrm{diag}(\delta_{l,i}(x))(\Theta_{l,i}+\rho I)^{-1},
\end{equation}
but it is not obvious whether we could obtain an analytical representation of this matrix.
It includes the summation over different $\delta_{l,i}(x')$ and $\delta_{l,i}(x)$. This makes the analysis much complicated.  At least, when $x'$ is given by the training samples, we can obtain the analytical formula as is shown in Lemma C.2.
Second, note that we have assumed $M_0\geq N$. When $M_0<N$, we have a singular $A_0$ and it makes the analysis more complicated. If we fix $W_1$ and train only the other weights $\{W_2$, ..., $W_L\}$, we can avoid the problem caused by the singular $A_0$ and achieve the fast convergence.

\subsection{Experiments}

\subsubsection{Setting of Figure 3}

 We computed condition numbers of various $\bar{\Theta}$ which were numerically obtained in
 a ReLU network with $L=3$ on synthetic data.
 We generated input samples $x$ by i.i.d. Gaussian, i.e., $x_i \sim \mathcal{N}(0,1)$. 
 We set, $C=1$, $M_0=100$, $N=80$, $\sigma_w^2=2$, $\sigma_b^2=0.5$ and $\rho=10^{-12}$.

\subsubsection{Fast convergence of unit-wise NGD}

\renewcommand{\thefigure}{S.1}
\begin{figure}
\vspace{-5pt}
\centering
\includegraphics[width=1\textwidth]{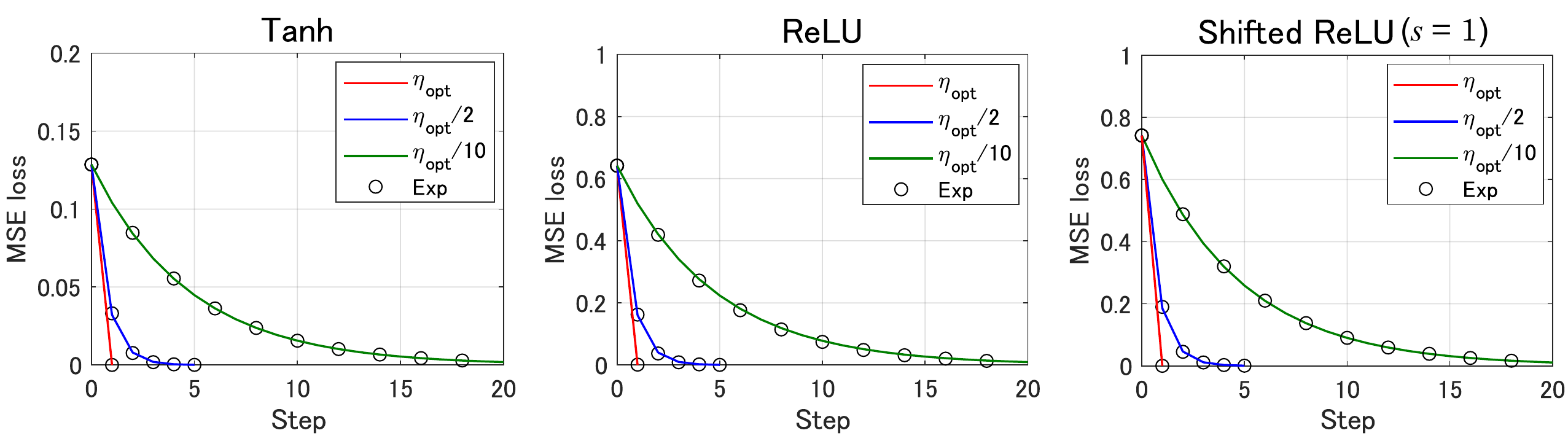}
\caption{Fast convergence of unit-wise NGD.
We trained deep networks with different activation functions ($L=3$, $C=1$, $M_l=M=4096$, $\sigma^2_w=2$, and $\sigma^2_b=0.5$) on two-class classification on MNIST ('0' and '7'; $N=100$). (Left) Tanh activation ($\alpha=M\times 2$). (Center) ReLU activation ($\alpha=M\times 1$). (Right) Shifted ReLU activation with $s=1$ ($\alpha=M\times 1.723...$).  }
\vspace{-5pt}
\end{figure}

 Figure S.1 shows an excellent agreement between our theory (given by Eq. (24); solid lines) and the experimental results of training (circles). In experiments, we used the unit-wise NGD, i.e., $G_{\text{unit},t}^{-1} \nabla_\theta \mathcal{L}$.
Depending on the activation function, we have different $\eta_{\text{opt}}=1/\alpha$. In the case of shifted ReLU, we used $\alpha$ obtained by using the analytical formula (\ref{S131:0611}).

\subsubsection{Check of \texorpdfstring{$\alpha$}{TEXT}}

Shifted ReLU activation is defined by  $\phi_s(x)=x \ (x\geq -s), \ -s \ (\text{otherwise})$. In this case,
Eq. (\ref{S131:0611}) becomes
\begin{equation}
    \alpha = \sum_{l=1}^{L-1} \left(\frac{1}{2} + \frac{1}{2} \mathrm{erf} (\frac{s}{\sqrt{2 q_l}})\right) M_l.
\end{equation}
In usual ReLU ($s=0$), we have $\alpha=\sum_{l=1}^{L-1}M_l/2$.

Figure S.2 shows that the above analytical values coincided well with numerical values (circles). We obtained the numerical values by directly computing the diagonal entries of $\bar{\Theta}$. We set 
$L=3$, $M_l=4096$, $M_0=N=10$, $\sigma^2_w=2$,  $\sigma^2_b=0.5$, and $\rho=10^{-12}$ to avoid numerical instability. We generated input samples $x$ by i.i.d. Gaussian, i.e., $x_i \sim \mathcal{N}(0,1)$.

\renewcommand{\thefigure}{S.2}
\begin{figure}
\vspace{-5pt}
\centering
\includegraphics[width=0.35\textwidth]{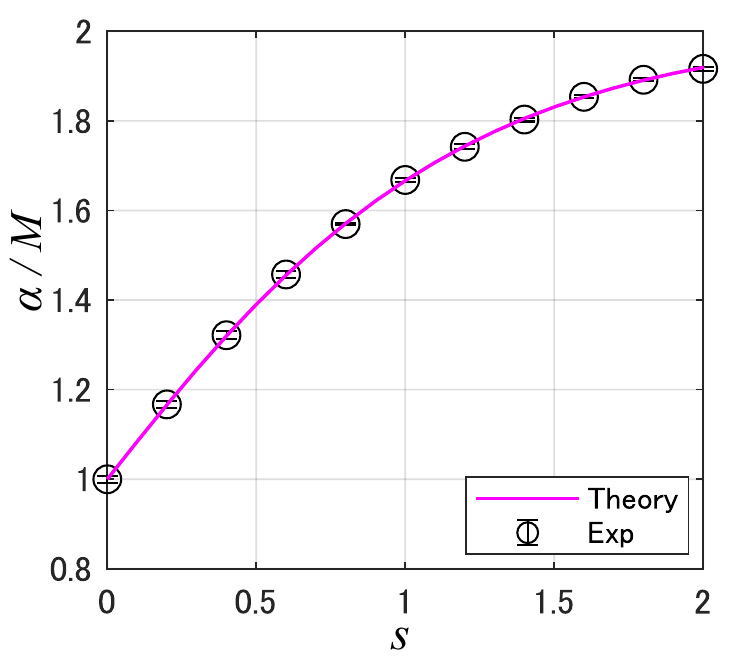}
\caption{$\alpha$ of networks with shifted ReLU $\phi_s$.}
\vspace{-5pt}
\end{figure}

\section{Fisher information for cross-entropy loss}

The FIM of the cross-entropy loss is known as  
\begin{equation}
G_t= \frac{1}{N} J^\top_t \Lambda(\sigma_t) J_t+\rho I,
\end{equation}
where $J_t=\nabla_\theta u_L$ is  Jacobian at time step $t$.  
  $\Lambda(\sigma)$ is  a block diagonal matrix which is composed $C \times C$ block matrices; $\mathrm{diag}(\sigma(x_n))-\sigma(x_n)\sigma(x_n)^\top$ ($n=1,...,N$) \cite{park2000,pascanu2013}.
 We denote softmax functions by $\sigma^{(k)}:=\exp(f_k)/\sum_{k'}^C \exp(f_{k'})$.
 Note that we always have $\Lambda_n 1_C= 0$ and $\Lambda(\sigma)$ is singular. The zero eigenvalue appears because $\sum_{k=1}^C \sigma^{(k)}(x_n) = 1$.
 This implies that a naive inversion of $G_t$  causes a gradient explosion. 
 To avoid the explosion, we add a damping term to $\Lambda$, such as $\Lambda+\tilde{\rho} I$. We have 
 \begin{equation}
  G_t= \frac{1}{N}J^\top_t (\Lambda(\sigma_t)+\tilde{\rho} I ) J_t+\rho I.
 \end{equation}

In the continuous time limit, exact NGD in the function space is given by 
\begin{align}
\frac{1}{\eta}\frac{d\sigma}{dt} &=\frac{1}{\eta} \frac{\partial \sigma}{\partial \theta} \frac{d\theta}{dt}\\ 
&= \Lambda_t J_t G^{-1}_t \nabla_\theta \mathcal{L} (\theta_t) \\
&= \Lambda_t \Theta_t ((\Lambda_t+\tilde{\rho}I) \Theta_t +\rho I)^{-1} (y-\sigma_t) \\ 
&= \Lambda_t (\Lambda_t+\tilde{\rho}I)^{-1}  (y-\sigma_t) \ \ \ \ \ \ \ \ (\rho=0), \label{S162:0610}
\end{align}
where we suppose that the NTK $\Theta_t$ is positive definite.  
Because $\Lambda_t$ includes $\sigma_t$, Eq. (\ref{S162:0610}) is a non-linear function of the softmax function. 
 It is not easy to explicitly solve the training dynamics even in the NTK regime (that is, $\Theta_t \sim \Theta_0$).

Next, we show that the above gradient  keeps unchanged even after taking the layer-wise approximation. We can consider the layer-wise approximation as 
\begin{equation}
    G_t= \frac{1}{N} S_t^\top(\Sigma \otimes (\Lambda_t+\tilde{\rho} I))S_t + \rho I,
\end{equation}
where $\Sigma$ is defined in the same way as in the FIM for the MSE loss. Then, we have the layer-wise NGD as 
\begin{align}
  \frac{1}{\eta} \frac{d\sigma}{dt}  
    &= \frac{1}{N}\Lambda_t J_t (\frac{1}{N}S_t^\top(\Sigma \otimes (\Lambda_t+\tilde{\rho} I))S_t + \rho I)^{-1} J_t^\top(y-\sigma_t) \\
    &= \frac{1}{N} \Lambda_t (1_L^\top \otimes I_{CN}) (S_tS_t^\top)(\frac{1}{N}\Sigma \otimes (\Lambda_t+\tilde{\rho} I))SS^\top + \rho I)^{-1} (1_L \otimes I_{CN}) (y-\sigma) \\ 
    &=   \Lambda_t ((1_L^\top \otimes I_{CN}) (\Sigma^{-1} \otimes (\Lambda_t+\tilde{\rho} I)^{-1}) (1_L \otimes I_{CN}) (y-\sigma_t) \ \ \ \ \ \ \ \ \ (\rho=0) \\
    &= \alpha \Lambda_t (\Lambda_t + \tilde{\rho} I)^{-1} (y-\sigma_t),
\end{align}
where $\alpha=1_L^\top \Sigma^{-1} 1_L$. 
Thus, the equation clarifies that we indeed obtain the same training dynamics as in the exact NGD by using layer-wise NGD with $\eta =c/\alpha$. The update in function space does not explicitly include NTK, as is the same as that for the MSE loss.

\section{Analytical kernels}
\label{secE:0416}

In this section, we summarize the analytical kernels that we used in numerical experiments.

The NTK is composed of an $N' \times N$ block matrix ($\Theta_{ana}$) \cite{jacot2018neural} such as 
\begin{equation}
    {\Theta}(x',x) = I_C \otimes  \Theta_{ana}(x',x)\com{/N},
\label{S148:0609}
\end{equation}
with
\begin{equation}
{\Theta}_{ana}(x',x) = \sigma_w^2  \sum_{l=1}^L B_l(x',x) \odot A_{l-1}(x',x) + \sigma_b^2  \sum_{l=1}^L  B_l(x',x).
\end{equation}
Each entries of feedforward signal block $A_l$ and feedback one $B_l$ are recursively computed as follows  \cite{karakida2018universal}: 
\begin{align}
A_l(x',x) &= \int Du_1 Du_2 \phi(\sqrt{q_l}u_1) \phi(\sqrt{q_l}( \bar{Q}_l(x',x) u_1+\sqrt{1-\bar{Q}_l(x',x)^2}u_2)), \label{anaA} \\
     B_l(x',x) &=  \sigma_w^2 \Xi_l(x',x) B_{l+1}(x',x), \label{anaB} \\ 
     \Xi_l(x',x)&= \int Du_1 Du_2 \phi'(\sqrt{q_l}u_1) \phi'(\sqrt{q_l}( \bar{Q}_l(x',x) u_1+\sqrt{1-\bar{Q}_l(x',x)^2}u_2)).
     \end{align}
 We denote an integral on Gaussian measure as $\int Du = \int du \exp(-u^2/2)/\sqrt{2 \pi}$.  
This analytical evaluation of the NTK is rigorously proved when the activation function is polynomially bounded \cite{yang2020tensor}.
 We have defined  
\begin{align}
 \bar{Q}_l (x',x) &:=Q_l(x',x)/q_l, \\ 
Q_l(x',x) &= \sigma_w^2 A_{l-1}(x',x) + \sigma_b^2, \\
q_l &:=  \sigma_w^2\int D u \phi(\sqrt{q_{l-1}}u)^2  +\sigma_b^2.
\end{align}
The scalar variable $q_l$ represents the amplitude of propagated signals. It is independent of $x$ because we normalize all of training and test samples by $\|x\|_2=1$ (that is, $q_0=1/M_0$). 
We can use the above $A_l$ and $B_l$ for
layer-wise NGD.  

For example, in ReLU networks, we have a matrix form of the kernels as follows:
\begin{align}
A_l(x',x) &= \frac{q_l}{2\pi}\left(  \sqrt{11^\top- \bar{Q}_l(x',x)^{\circ 2}}+\frac{\pi}{2} \bar{Q}_l(x',x)+\bar{Q}_l(x',x) \odot \arcsin (\bar{Q}_l(x',x)) \right),  \\
     \Xi_l(x',x)&= \frac{\com{1}}{2\pi}  \left(\arcsin ( \bar{Q}_l(x',x)) + \frac{\pi}{2} 11^\top \right), \\
     q_l &=  \frac{\sigma_w^2}{2}q_{l-1} +\sigma_b^2 \ \ \ \ (l\geq2),\ \ q_1 = \sigma^2/M_0 + \sigma_b^2,
     \end{align}
where $(\cdot)^{\circ 2}$ means entry-wise square.

\end{document}